\documentclass[journal]{IEEEtran}
\usepackage{color}
\usepackage{bm}
\usepackage{amsfonts}
\usepackage{algpseudocode}
\usepackage{algorithmicx,algorithm}
\usepackage{amsmath}
\usepackage{multirow}
\usepackage{subfig}
\usepackage{ulem}

\ifCLASSOPTIONcompsoc
  \usepackage[nocompress]{cite}
\else
  \usepackage{cite}
\fi

\ifCLASSINFOpdf
   \usepackage[pdftex]{graphicx}
\else
\fi

\usepackage{stfloats}
\begin{document}
%
\title{Joint Deep Learning of Facial Expression Synthesis and Recognition}
%
%
%
%

\author{Yan~Yan,~\IEEEmembership{Member,~IEEE,}
              Ying~Huang,
              Si~Chen,
              Chunhua~Shen
              and Hanzi~Wang,~\IEEEmembership{Senior Member,~IEEE,}
\IEEEcompsocitemizethanks{\IEEEcompsocthanksitem Y.~Yan, Y.~Huang, H.~Wang are with the Fujian Key Laboratory of Sensing and Computing for Smart City, School of Informatics,
Xiamen University, Xiamen 361005, China
(email: yanyan@xmu.edu.cn; ying\_hwang@qq.com; hanzi.wang@xmu.edu.cn).\protect
\IEEEcompsocthanksitem S.~Chen is with School of Computer and Information Engineering, Xiamen University of Technology, Xiamen 361024, China (email: chensi@xmut.edu.cn).\protect
\IEEEcompsocthanksitem C.~Shen is with School of Computer Science, The University of Adelaide, Adelaide, SA 5005, Australia (e-mail: chunhua.shen@adelaide.edu.au).\protect
}
}

%
%

\markboth{}
{Shell \MakeLowercase{\textit{et al.}}: Bare Demo of IEEEtran.cls for Computer Society Journals}
%



\IEEEtitleabstractindextext{%
\begin{abstract}
Recently, deep learning based facial expression recognition (FER) methods have attracted considerable attention and they usually require large-scale labelled training data. Nonetheless, the publicly available facial expression databases typically contain a small amount of labelled data. In this paper, to overcome the above issue, we propose a novel joint deep learning of facial expression synthesis and recognition method for effective FER. More specifically, the proposed method involves a two-stage learning procedure. Firstly, a facial expression synthesis generative adversarial network (FESGAN) is pre-trained to generate facial images with different facial expressions. To increase the diversity of the training images, FESGAN is elaborately designed to generate images with new identities from a prior distribution. Secondly, an expression recognition network is jointly learned with the pre-trained FESGAN in a unified framework. In particular, the classification loss computed from the recognition network is used to simultaneously optimize the performance of both the recognition network and the generator of FESGAN. Moreover, in order to alleviate the problem of data bias between the real images and the synthetic images, we propose an intra-class loss with a novel real data-guided back-propagation (RDBP) algorithm to reduce the intra-class variations of images from the same class, which can significantly improve the final performance. Extensive experimental results on public facial expression databases demonstrate the superiority of the proposed method compared with several state-of-the-art FER methods.
\end{abstract}

\begin{IEEEkeywords}
Facial expression recognition, facial expression synthesis, convolutional neural networks (CNNs), generative adversarial net (GAN).
\end{IEEEkeywords}}

\maketitle

\IEEEdisplaynontitleabstractindextext

%
\IEEEpeerreviewmaketitle

\section{Introduction}\label{sec:introduction}

\IEEEPARstart{O}{ver} the past few decades, facial expression recognition (FER) has attracted considerable attention in computer vision \cite{haxby2000distributed, corneanu2016survey}, due to its practical importance in a variety of applications, including sociable robotics, health care and human-computer interaction \cite{martinez2016advances}. Despite significant progress has been made \cite{zhong2012learning,liu2014feature,liu2014facial,xie2019facial}, FER remains a great challenge, which mainly suffers from variations in facial appearance caused by different poses, changing illumination and partial occlusions.

In recent years, deep learning has achieved outstanding performance in many computer vision tasks, including image classification \cite{he2016deep}, face recognition \cite{tran2017disentangled}. The great success of deep learning is mainly due to the powerful representation capability of deep neural networks and the available large-scale labelled training data. Until recently, deep learning based FER methods \cite{mollahosseini2016going,ng2015deep,ding2017facenet2expnet} have also been developed. However, the performance of these FER methods under unconstrained environments is still far from being satisfactory. One reason is that the publicly available facial expression databases typically contain a small number of training data. Although rich and diverse facial images are available on the Internet, manual labelling of these images is time-consuming and expensive. As a result, it is not trivial to train a deep neural network using a limited number of training data.

\begin{figure}[!t]
\centering
\includegraphics[width=2.6in]{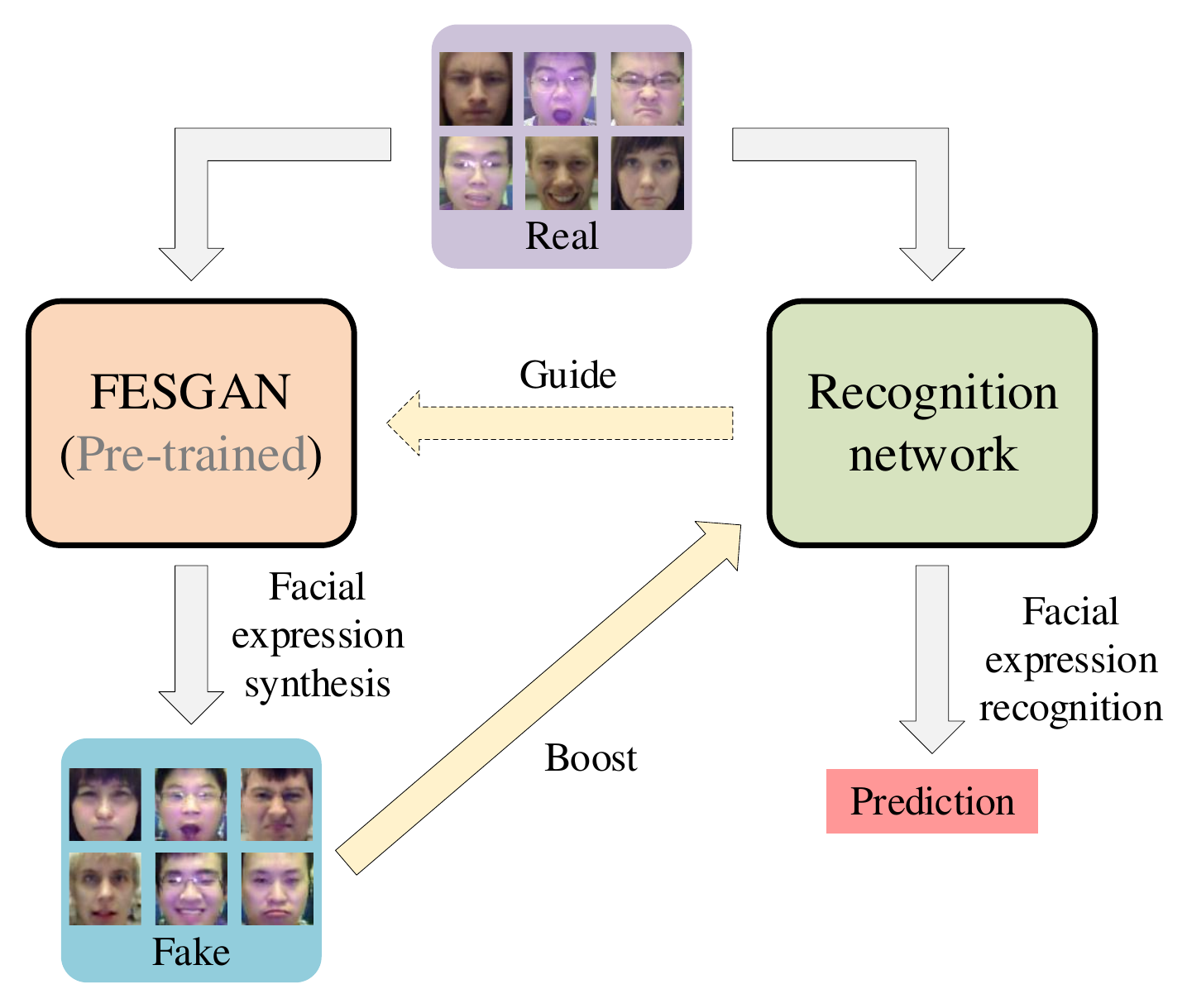}
\caption{\small Overview of the proposed method. FESGAN and the recognition network are jointly optimized, where the performance of the recognition network is boosted by the outputs from FESGAN, and the optimization of FESGAN is guided by the recognition network for realistic image synthesis.}
\label{fig:illustration}
\end{figure}

To alleviate the problem caused by the lack of insufficient training data, some works attempt to take advantage of auxiliary data to train the neural networks for effective FER. For example, a deep convolutional neural network (CNN) architecture \cite{mollahosseini2016going} trained on the hybrid databases (consisting of seven different facial expression databases) is proposed to achieve comparable results on each database. However, due to the bias between these databases \cite{li2018deep}, such a training strategy may lead to the problem of overfitting and thus degrade the performance on the target database. In \cite{ng2015deep,ding2017facenet2expnet}, a CNN is firstly pre-trained on a large-scale face recognition database or other large-scale image databases (such as Imagenet \cite{deng2009imagenet} and VGGFace \cite{parkhi2015deep}), and then fine-tuned on the target facial expression database. Nevertheless, designing an appropriate strategy to perform fine-tuning usually requires considerable efforts, since the deep networks are pre-trained with high capacity for large-scale data. Although some tricks (such as simple data augmentation and dropout \cite{srivastava2014dropout}) show the advantage to prevent the CNN from overfitting, the performance improvements are limited. Therefore, how to effectively improve the performance of FER under insufficient training data is still an open problem.

Recently, a new generative model, termed generative adversarial network (GAN) \cite{goodfellow2014generative}, has received great interests due to its promising performance to generate high-quality data. GAN usually consists of a generator network and a discriminator network. Through the adversarial optimization between these two networks, GAN is able to generate synthetic samples that simulate the distribution of real samples from training data. Particularly, GAN has been successfully used in some face-related tasks, such as posed face synthesis \cite{tran2017disentangled} and facial attributes transfer \cite{choi2018stargan}. These methods are able to generate photo-realistic facial images having the same identity as the input facial images. Due to the similarities between the synthetic images and the real images, these synthetic images that vary in different conditions (e.g., expressions, poses) can be used for data augmentation to alleviate the problem of insufficient training data in deep learning.

However, there are still several unsolved problems when these synthetic images are directly used to train a deep neural network. For example, generating high-quality images that approximate the distribution of real images, usually requires considerable efforts for GAN, especially when limited training data are available. In addition, although the synthetic images generated by GAN are of high-quality (even though they cannot be distinguished by human eyes), it cannot guarantee that the performance of deep neural network can be effectively improved. This is because that the intrinsic data bias between the synthetic images and the real images may be large for the recognition network.
%
%

In this paper, we propose a novel joint deep learning of facial expression synthesis and recognition method for effective FER. In the proposed method, the tasks of facial expression synthesis and recognition are simultaneously performed in a unified framework to boost the performance of each task. The proposed method involves a two-stage learning procedure (see Fig. \ref{fig:illustration}). Firstly, a facial expression synthesis GAN (FESGAN) is pre-trained to generate synthetic facial images with different facial expressions. Secondly, a recognition network is jointly trained with the pre-trained FESGAN in a unified framework.

In summary, the main contributions of this paper are summarized as follows:

\begin{itemize}
\item A FESGAN and a recognition network are integrated in a unified framework to perform facial expression synthesis and recognition simultaneously. The classification loss obtained from the recognition network for the synthetic images is also responsible for the training of generator in FESGAN. As a result, the synthetic facial images generated from FESGAN are beneficial to the training of the recognition network.
\item An intra-class loss is introduced to mitigate the problem of data bias between the real facial images and the synthetic facial images in the recognition network. In this manner, we not only largely overcome the problem of overfitting, but also enhance the intra-class compactness for obtaining features with high discriminability.
\item A novel real data-guided back-propagation (RDBP) algorithm is developed to optimize the intra-class loss. RDBP takes advantage of the features of the real facial images to supervise the learning of the features of the synthetic facial images from the same class. Therefore, it can effectively avoid degrading the recognition capability for the real facial images due to interference from the synthetic facial images.
\end{itemize}

The remaining part of this paper is organized as follows. Section 2 reviews the related work. Section 3 firstly gives an overview of the proposed method and then describes the details of facial expression synthesis GAN (FESGAN) and the recognition network, respectively. Section 4 presents the experimental results. Finally, Section 5 gives the final conclusion.

\section{Related work}
In this section, we briefly review some related work on GAN, facial expression synthesis and facial expression recognition, respectively.

\subsection{Generative Adversarial Network (GAN)}

Generative adversarial network (GAN) \cite{goodfellow2014generative} is a powerful framework to learn the generative model, where a minimax two-player game strategy is adopted to optimize a generator and a discriminator. Through the game, the generator aims to maximally fool the discriminator, while simultaneously the discriminator seeks to discriminate the generated samples from the true ones as much as possible. In this manner, the performance of both the generator and the discriminator can be improved.

A variety of GAN-based methods have been proposed for different computer vision tasks, such as image synthesis \cite{radford2015unsupervised}, image super-resolution \cite{ledig2017photo}, image style transfer \cite{zhu2017unpaired}. These methods are proposed to improve the original GAN from different perspectives. One crucial extension of GAN is conditional GAN (cGAN) \cite{mirza2014conditional}, where a conditional model is developed to guide the generation of images by conditioning the input on the specific category. In InfoGAN \cite{chen2016infogan}, a variational regularization term is introduced into the optimization of GAN, where the mutual information between the latent variables and the observations of the generator is expected to be maximized. In CycleGAN \cite{zhu2017unpaired}, a cycle structure is employed to perform unpaired image-to-image translation, and a cycle consistency loss is developed to preserve the content of the original images while only changing the style of the inputs.

\subsection{Facial Expression Synthesis}
Facial expression synthesis has been an active research area in computer vision and graphics. A large number of facial expression synthesis methods have been developed over the past few decades \cite{Testa2019}. Traditional methods include the 3D model-based methods \cite{Blanz2003}, the flow-based methods \cite{Yang2011} and the 2D expression mapping-based methods \cite{Liu2001,Zhang2005,Xie2017}. Recently, the deep learning-based methods have been proposed.  In the following, we mainly review the deep learning-based methods.

%
%
%

Susskind et al. \cite{Susskind2008} develop a deep belief network (DBN) method, which makes use of the high-level descriptors (including facial action coding system (FACS) labels and
identity labels) to generate facial expression images.
Reed et al. \cite{Reed2014} propose a higher-order Boltzman machine to model the distribution over features and the latent factors of variation (such as pose, expression). 
However, the resolution of the generated images is relatively low ($48\times48$) and these generated images tends to be blurry.
 Yeh et al. \cite{Yeh2016} train a flow variational autoencoder, which is able to synthesize facial images by taking advantage of the flow-based face manipulation. But this method requires facial images taken under similar lighting conditions and background as the training data, due to the use of $l_2$ norm.
 Li et al. \cite{Li2016} propose a deep identity-aware transfer (DIAT) model for facial attribute transfer given the source input image and the reference attribute.
Recently, Chang et al. \cite{Chang2019} propose a deep learning-based method for face alignment, modeling, and expression estimation (FAME), which regresses the 3D morphable shape, 3D expression and viewpoint parameters directly from the images.

With the development of GAN, it has been successfully applied to the task of facial expression synthesis. For example, in \cite{choi2018stargan}, StarGAN is proposed to address multi-domain image-to-image translation across different databases. To train the model on the unlabelled database, a mask vector is introduced to ignore the unlabelled categories so that the model can focus on exploiting the labelled database. Thus, an unlabelled facial image can transfer to the image with a given facial expression label while preserving the information of other domains. In \cite{song2018geometry}, a geometry-guided GAN (G2-GAN) is proposed to take advantage of facial geometry information to guide facial expression synthesis. In G2-GAN, a pair of GANs is employed to simultaneously perform facial expression removal and synthesis. In \cite{qiao2018geometry}, a geometry-contrastive GAN (GC-GAN) is proposed to  synthesize facial expression images conditioned on geometry information at the semantic level. Different from G2-GAN, which treats the facial landmarks as an additional image channel concatenated with the input image, GC-GAN employs a network that is specifically designed to perform facial geometry embedding by contrastive learning.
In \cite{ding2018exprgan}, an expression GAN (ExprGAN) for photo-realistic
facial expression editing with controllable expression intensity is proposed for facial expression synthesis.  In \cite{Bozorgtabar2019}, a new attribute guided facial image synthesis based on GAN is developed to perform image-to-image translation.

\subsection{Facial Expression Recognition (FER)}


Traditional FER methods usually employ hand-crafted features, such as local binary patterns (LBP) \cite{shan2009facial}, local phase quantization (LPQ) \cite{tawari2013face}, histograms of oriented gradients (HOG) \cite{hu2008multi} and scale invariant feature transform (SIFT) \cite{zhang2016deep}. In \cite{zhong2012learning}, Zhong et al.~show that only a few face patches play an important role in FER, and these patches can be grouped into common patches (which are important for all facial expressions) and specific patches (which are only active to specific facial expressions). Therefore, a two-stage multi-task sparse learning is proposed to select these patches for FER. However, the common patches and specific patches may be overlapped, since these two groups of patches are independently learned. As a result, some redundant information may be preserved, which decreases the discriminability of the final features. To solve this problem, Liu et al. \cite{liu2014feature} propose a unified framework, where the sparse vector machine and multi-task learning are combined to select the specific features and common features. Note that these methods consider feature learning and classifier training as two separate steps, which may lead to suboptimal performance.

Recently, due to the remarkable success achieved by deep learning in various computer vision tasks \cite{zhang2016joint,ren2017faster,liu2017sphereface}, some studies begin to exploit deep learning for FER. Liu et al. \cite{liu2014facial} employ a boosted deep belief network (DBN) based framework to jointly perform feature learning, feature selection and classifier construction. In \cite{jung2015joint}, Jung et al. firstly train two independent deep neural networks to learn temporal appearance features and temporal geometry features, respectively. Then, these two networks are jointly fine-tuned to boost the performance of FER. In \cite{zhao2016peak}, Zhao et al. propose a peak-piloted deep network, where the inputs of the deep network are the peak and non-peak facial expression images from the same class and the same identity. To learn intensity-invariant facial expression features, the L2-norm loss is used to reduce the distances of features between the peak facial expression and the non-peak facial expression. 
\section{The proposed method}
\subsection{Overview}

\begin{figure*}[!t]
\centering
\includegraphics[width=6.6in]{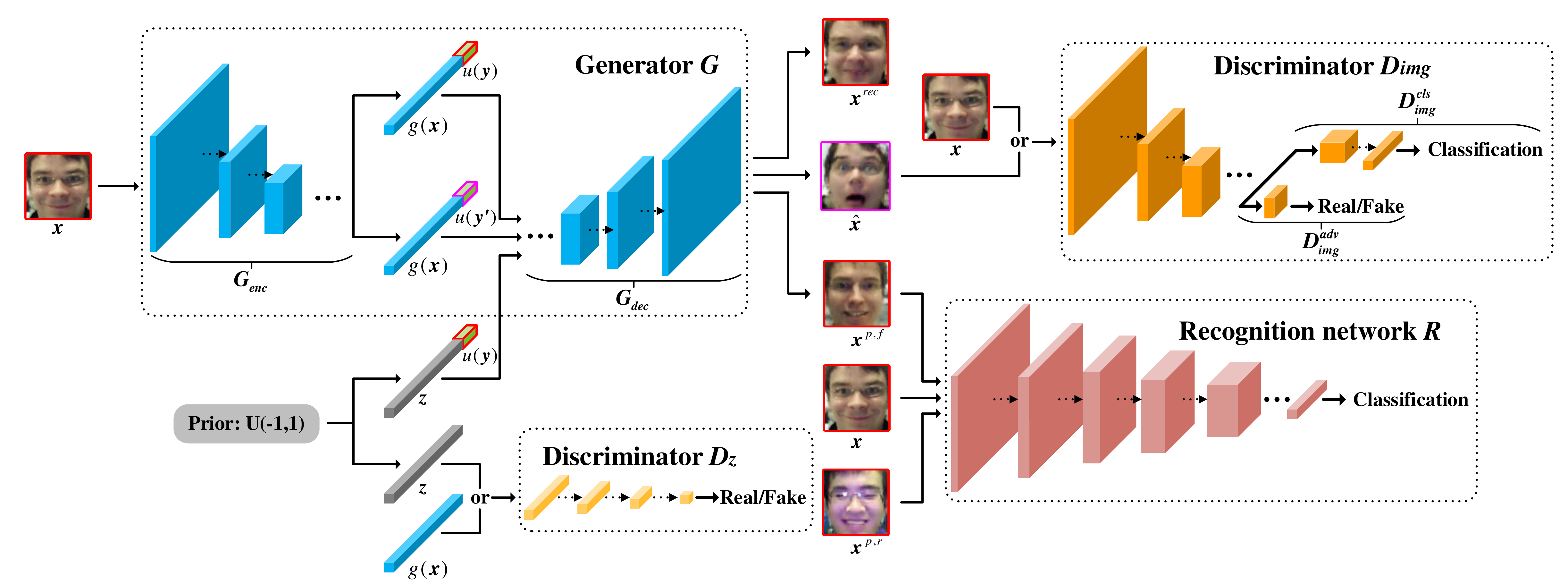}
\caption{\small The overview of our proposed method.}
\label{fig:overview}
\end{figure*}

In this paper, we propose a novel joint deep learning of facial expression synthesis and recognition method (called FESR) for effective facial expression recognition.
The overview of the proposed method is given in Fig. \ref{fig:overview}, which involves a two-stage learning procedure.

In the first stage, a facial expression synthesis GAN (FESGAN) consisting of a generator $G$ and two discriminators (i.e., $D_{img}$ and $D_{z}$) is pre-trained to generate synthetic facial images. In this paper,
the autoencoder structure (including an encoder $G_{enc}$ and a decoder $G_{dec}$) \cite{zhang2017age} is employed as the generator. The generator of FESGAN is used to learn the content style of  real facial images. Thus, the adversarial learning between the generator and the discriminators, and the content learning between the real images and the generated images are simultaneously performed to enhance the quality of the generated facial images.

More specifically, the encoder $G_{enc}$ firstly maps the input facial image $\bm{x}$ into a latent face representation $g(\bm{x})$. Then, the decoder $G_{dec}$ takes both the face representation $g(\bm{x})$ and the continuous representation of a given label (which effectively depicts both the category and intensity of facial expression) as the inputs to generate the facial image $\bm{\hat{x}}$. The generated facial image $\bm{\hat{x}}$ is expected to be photo-realistic, which not only captures the facial expression information, but also preserves the identity information of the input image $\bm{x}$. Finally, a discriminator (i.e., $D_{img}$) is adopted to distinguish the real images from the synthetic ones while performing FER.
Meanwhile, another discriminator (i.e., $D_{z}$) is used to discriminate the latent face representation $g(\bm{x})$ and the random vector $\bm{z}$ sampled from a prior distribution. Therefore, FESGAN is able to generate facial images from a prior distribution (these images are used as the auxiliary training data for the recognition network $R$)


In the second stage, a recognition network $R$ is introduced to be jointly trained with the pre-trained FESGAN in a unified framework. The recognition network $R$ takes both the real facial images ($\bm{x}$ and $\bm{x}^{p,r}$) and the synthetic facial image ($\bm{x}^{p,f}$) generated by FESGAN as the inputs to perform facial expression classification. Meanwhile, the classification loss obtained from the recognition network for the synthetic images is also used to train the generator of FESGAN. In this manner, FESGAN can generate facial images that are beneficial to the training of the recognition network. Furthermore, to mitigate the influence of data bias between the real images and the synthetic images, a novel intra-class loss is introduced to reduce the intra-class variations of both the real and synthetic images. In particular, a specifically designed real data-guided back-propagation (RDBP) algorithm is developed. RDBP takes advantage of the features of the real facial images to supervise the learning of the features of the synthetic facial images from the same class. As a result, the recognition network is able to make full use of facial expression information from both the real images and the synthetic images, thus significantly improving the performance of FER.

In the following subsections, we will respectively introduce FESGAN, the recognition network and the training algorithm of the proposed method in detail.

\subsection{Facial Expression Synthesis GAN (FESGAN)}
\label{sec:fesgan}

The architecture of the proposed FESGAN is shown in Fig. \ref{fig:overview}. In general, FESGAN contains three components: a generator $G$ and two discriminators (i.e., $D_{img}$ and $D_{z}$). The generator $G$ has two parts: the encoder $G_{enc}$ and the decoder $G_{dec}$. The discriminator $D_{img}$ gives two outputs, where one output $D_{img}^{adv}$ determines whether the input image is `real' or `fake' (synthetic) by adversarial learning, while the other output $D_{img}^{cls}$ predicts the expression category of the input image by classification. Another discriminator $D_{z}$ determines whether the input is the latent face representation or the random vector sampled from a prior distribution.

Specifically, given an input facial image $\bm{x}$ with the label $\bm{y}=(y_{1},...,y_{K})$, where $K$ is the number of facial expressions, and $y_{i=m}=1$, $y_{j\neq m}=0$ indicates that $\bm{x}$ belongs to the $m_{th}$ facial expression. The encoder $G_{enc}$, which consists of several convolutional layers and one fully-connected layer, is used to map the input image $\bm{x}$ into a latent face representation, denoted as,
\begin{equation}
\label{eq:gx}
g(\bm{x})=G_{enc}(\bm{x}),
\end{equation}
where $g(\bm{x})\in\mathbb{R}^{n}$, and $n$ is the dimension of the face representation space.

Given another label $\bm{y'}$ (which also depicts the category information of facial expression and is different from the expression label $\bm{y}$), the decoder is expected to synthesize a facial image that is conditioned on $\bm{y'}$ and the latent representation $g(\bm{x})$.
Note that, each facial expression usually appears in either a subtle form (i.e., non-peak expression) or a strong form (i.e., peak expression). Therefore, in order to manipulate the intensity of the facial expression and capture the critical and subtle expression details, we firstly transform the label $\bm{y'}$ to a continuous representation as follows \cite{he2017arbitrary,ding2018exprgan},
\begin{equation}
\bm{u}(\bm{y'})=\bm{v}\odot(2\bm{y'}-1),
\end{equation}
where $\bm{v} = (v_1,...,v_K)$ has the same dimensionality as the label $\bm{y'}$ and each element of $\bm{v}$ is subject to a uniform distribution on [0,1] (i.e., $v_{i}\sim U(0,1)$). Therefore, the output ${u}(\bm{y'})=(u(y'_{1}),...,u(y'_{K}))$ is continuous and $u(y'_{i})\sim U(-1,1)$. Obviously, $u(y'_{i})$ effectively encodes the intensity of the $i_{th}$ facial expression.

Then, the decoder $G_{dec}$, which consists of several transposed convolutional layers, takes the concatenation of the face representation $g(\bm{x})$ and the continuous representation $u(\bm{y'})$ as the input to generate the synthetic image $\bm{\hat{x}}$, denoted as,
\begin{equation}
\label{eq:xb}
\bm{\hat{x}}=G_{dec}([g(\bm{x}), u(\bm{y'})]).
\end{equation}
where $[g(\bm{x}), u(\bm{y'})]$ denotes the concatenation of $g(\bm{x})$ and $u(\bm{y'})$.
 In this way, $G_{dec}$ is able to synthesize the facial image with a given facial expression and controllable expression intensity.

In the following, we will introduce the objective functions for the training of FESGAN in detail. 

\textbf{The adversarial loss.} To generate the photo-realistic synthetic image $\bm{\hat{x}}$, adversarial learning is adopted between the generator $G$ and the discriminator $D_{img}$, which can be used to classify the `real'  and `fake' (synthetic) images. Inspired by Wasserstein GAN \cite{arjovsky2017wasserstein}, the adversarial losses for $G$ and $D_{img}$ are respectively formulated as,
\begin{equation}
\mathcal{L}_{adv_{g}}^{img} =-\mathbb{E}_{\bm{x},\bm{y'}}[D_{img}^{adv}(\bm{\hat{x}})],
\end{equation}
\begin{equation}
\mathcal{L}_{adv_{d}}^{img} =-\mathbb{E}_{\bm{x}}[D_{img}^{adv}(\bm{x})]+\mathbb{E}_{\bm{x},\bm{y'}}[D_{img}^{adv}(\bm{\hat{x}})],
\end{equation}
where $D_{img}^{adv}(\bm{x})$ and $D_{img}^{adv}(\bm{\hat{x}})$ represent the outputs of $D_{img}$ (composed of several convolutional layers) for the real facial image $\bm{x}$ and the synthetic facial image $\bm{\hat{x}}$, respectively. The adversarial learning between $G$ and $D_{img}$ is beneficial to generate photo-realistic facial images.

In order to increase the diversity of the generated facial images, we impose a prior distribution (the uniform distribution is used in this paper) on the latent face representation $g(\bm{x})$ by taking advantage of another discriminator $D_{z}$ (composed of several fully-connected layers). The adversarial losses for $G_{enc}$ and $D_{z}$ are respectively defined as,
\begin{equation}
\mathcal{L}_{adv_{g}}^{z} =-\mathbb{E}_{\bm{x}}[\log(D_{z}(g(\bm{x}))],
\end{equation}
\begin{equation}
\mathcal{L}_{adv_{d}}^{z} =-\mathbb{E}_{\bm{z}}[\log D_{z}(\bm{z})]-\mathbb{E}_{\bm{x}}[\log(1-D_{z}(g(\bm{x})))],
\end{equation}
where $\bm{z}=(z_1,...,z_n)$  is a random vector sampled from the prior distribution (i.e., $z_{i}\sim U(-1,1)$).

The above adversarial losses force the face representation to populate the latent space evenly with no apparent `holes' (see \cite{zhang2017age} for more details). Therefore, the decoder $G_{dec}$ is able to generate facial images with new identities by using $\bm{z}$ instead of $g(\bm{x})$, which can effectively approximate the appearance of the real face and do not deviate from the face manifold.
In other words, the adversarial learning between $G_{dec}$ and $D_z$ guarantees smooth transition in the face representation space.

Note that these two discriminators (i.e., $D_{img}$ and $D_{z}$) used in FESGAN empirically employ different forms of adversarial losses. This is mainly because that the adversarial losses of Wasserstein GAN are more suitable for dealing with the image as the input as observed in \cite{arjovsky2017wasserstein}, while those of the original GAN can better handle the vector as the input.

\textbf{The classification loss.} For facial expression synthesis, the synthetic image generated by the generator $G$ should be capable of showing the corresponding facial expression given the label of expression $\bm{y'}$. To achieve this, an auxiliary classifier is added in the discriminator $D_{img}$ to perform FER. Note that, the facial expression of real facial image should also be correctly recognized by the classifier. Therefore, the classification loss for the synthetic and real images in the auxiliary classifier consist of two terms to optimize $G$ and $D_{img}$, respectively.

On one hand, the classification loss for the synthetic image $\bm{\hat{x}}$, which is used to optimize $G$, is defined as
\begin{equation}
\mathcal{L}_{cls,D}^{f}= -\mathbb{E}_{\bm{x},\bm{y'}}[\log D_{img}^{cls}(\bm{y'}|\bm{\hat{x})}].
\end{equation}
where $D_{img}^{cls}(\bm{y'}|\bm{\hat{x}})$ denotes the probability of correct prediction for $\bm{\hat{x}}$. The above classification loss facilitates the generator to enhance its ability of synthesizing visually photo-realistic facial expressions.

On the other hand, the classification loss for the real facial image $\bm{x}$, which is used to optimize $D_{img}$, is defined as,
\begin{equation}
\mathcal{L}_{cls,D}^{r}= -\mathbb{E}_{\bm{x},\bm{y}}[\log D_{img}^{cls}(\bm{y}|\bm{x})],
\end{equation}
where $D_{img}^{cls}(\bm{y}|\bm{x})$ denotes the probability of correct prediction for $\bm{x}$. Such a loss encourages the discriminator to correctly recognize the facial expression.

In general, by minimizing $\mathcal{L}_{cls,D}^{f}$, $G$ tries to generate the synthetic images that can be correctly classified by $D_{img}$. Meanwhile, by minimizing $\mathcal{L}_{cls,D}^{r}$, $D_{img}$ also learns to classify the real images as the correct class.

\textbf{Content learning.} The synthetic image generated from the generator $G$ is supposed to not only fool the discriminator $D_{img}$, but also it preserves the content of the input image. To achieve this, we employ two pixel-level losses (i.e., the reconstruction loss $\mathcal{L}_{rec}$ and the identity preserving loss $\mathcal{L}_{id}$) to encourage the generator $G$ to effectively learn the content from the real facial images.

For the reconstruction loss $\mathcal{L}_{rec}$, the generator $G$ is required to be capable of reconstructing the input image $\bm{x}$ so as to preserve the content of the input image, which is defined as,
\begin{equation}
\mathcal{L}_{rec}=\mathbb{E}_{\bm{x},\bm{y}}[\left \| \bm{x}-\bm{x}^{rec} \right \|_{1}],
\end{equation}
where $\bm{x}^{rec}$ denotes the reconstructed image, that is,
\begin{equation}
\label{eq:xrec}
\bm{x^}{rec}=G_{dec}([g(\bm{x}),u(\bm{y})]).
\end{equation}

To preserve the identity information of the input image, we also employ the identity preserving loss to keep identity consistency, which is defined as,
\begin{equation}
\mathcal{L}_{id}=\mathbb{E}_{\bm{x},\bm{y}}[\left \| F_{id}(\bm{x}^{rec})-F_{id}(\bm{x}) \right \|_{1}],
\end{equation}
where $F_{id}$ is a feature extractor for facial identity. In this paper, we employ the Light CNN-29 \cite{wu2018light} for feature extraction, which is pre-trained on large-scale face databases for face recognition.

By the content learning, the generator $G$ is able to learn the content style of the input facial images. Therefore, the generator can output the facial images having the similar content style as the input facial images.

In this paper, we take advantage of the adversarial loss and the discriminator $D_z$, which can force the face representation to obey the prior distribution (i.e., the uniform distribution in this paper). In this manner, the generator is able to generate  facial images with new identities by using the random vector.
It is worth pointing out that the Kullback-Leibler (K-L) loss can also be used to reduce the gap between the prior distribution ($\bm{z}$) and the latent distribution ($g(\bm{x})$). In fact, the K-L loss plays the role of a regularizer (which is widely used in Variational Autoencoder (VAE) \cite{Kingma2014}, which regularizes the encoder by imposing a prior over the latent distribution). However, the K-L loss cannot guarantee that the generator can output photo-realistic facial images by using the random vector (note that the optimization problem of the K-L loss only considers the mean and variance of the latent vector). Moreover, the results from VAE are usually blurry \cite{Bao2017}.
In contrast, the discriminator $D_z$ is explicitly trained to distinguish the random vector sampled from the prior distribution from the encoded latent representation ($g(\bm{x})$). Therefore, based on the adversarial loss, the generator is able to synthesize photo-realistic facial images.
\subsection{The Recognition Network}

The recognition network is denoted as $R$, which consists of a feature extractor $R_{ext}$ (composed of several convolutional layers and one fully-connected layer) and a classifier $R_{cls}$ (composed of one fully-connected layer). In this paper, $R$ takes a triplet (i.e., $\{\bm{x},\bm{x}^{p,r},\bm{x}^{p,f}\}$) as the input, where these three images are from the same class (belonging to the same facial expression). $\bm{x}$ represents one real facial image taken from the training set.
$\bm{x}^{p,r}$ denotes another real facial image from the same class as $\bm{x}$, and $\bm{x}^{p,f}$ denotes the synthetic image generated by $G_{dec}$, which is defined as,
\begin{equation}
\label{eq:xpf}
\bm{x}^{p,f}=G_{dec}([\bm{z},u(\bm{y})]).
\end{equation}
Based on the triplet input, a novel intra-class loss and a classification loss are adopted to supervise the training of $R$. 

\textbf{The intra-class loss.} As we discuss in Section \ref{sec:introduction}, it cannot be guaranteed that the performance of the recognition network can be effectively improved  by taking advantage of the synthetic images generated by FESGAN. The intrinsic data bias between the real images and the generated images may be large, which can degrade the final performance of FER. To mitigate this, we introduce an intra-class loss to reduce the intra-class variations of images from the same class, which is formulated as,
\begin{equation}
\mathcal{L}_{intra}=\mathcal{L}_{dist}^{r}(\bm{x},\bm{x}^{p,r})+\mathcal{L}_{dist}^{rf}(\bm{x},\bm{x}^{p,f}),
\end{equation}
where $\mathcal{L}_{dist}^{r}(\bm{x},\bm{x}^{p,r})$ is the Euclidean distance between the features of $\bm{x}$ and $\bm{x}^{p,r}$, and $\mathcal{L}_{dist}^{rf}(\bm{x},\bm{x}^{p,f})$ is the Euclidean distance between the features of $\bm{x}$ and $\bm{x}^{p,f}$, that is,
\begin{equation}
\mathcal{L}_{dist}^{r}(\bm{x},\bm{x}^{p,r})=\left \| R_{ext}(\bm{x})-R_{ext}(\bm{x}^{p,r}) \right \|_{2},
\end{equation}
\begin{equation}
\mathcal{L}_{dist}^{rf}(\bm{x},\bm{x}^{p,f})=\left \| R_{ext}(\bm{x})-R_{ext}(\bm{x}^{p,f}) \right \|_{2}.
\end{equation}
Here, $R_{ext}(.)$ denotes the feature extractor.

\begin{figure}[!t]
\centering
\includegraphics[width=3.1in]{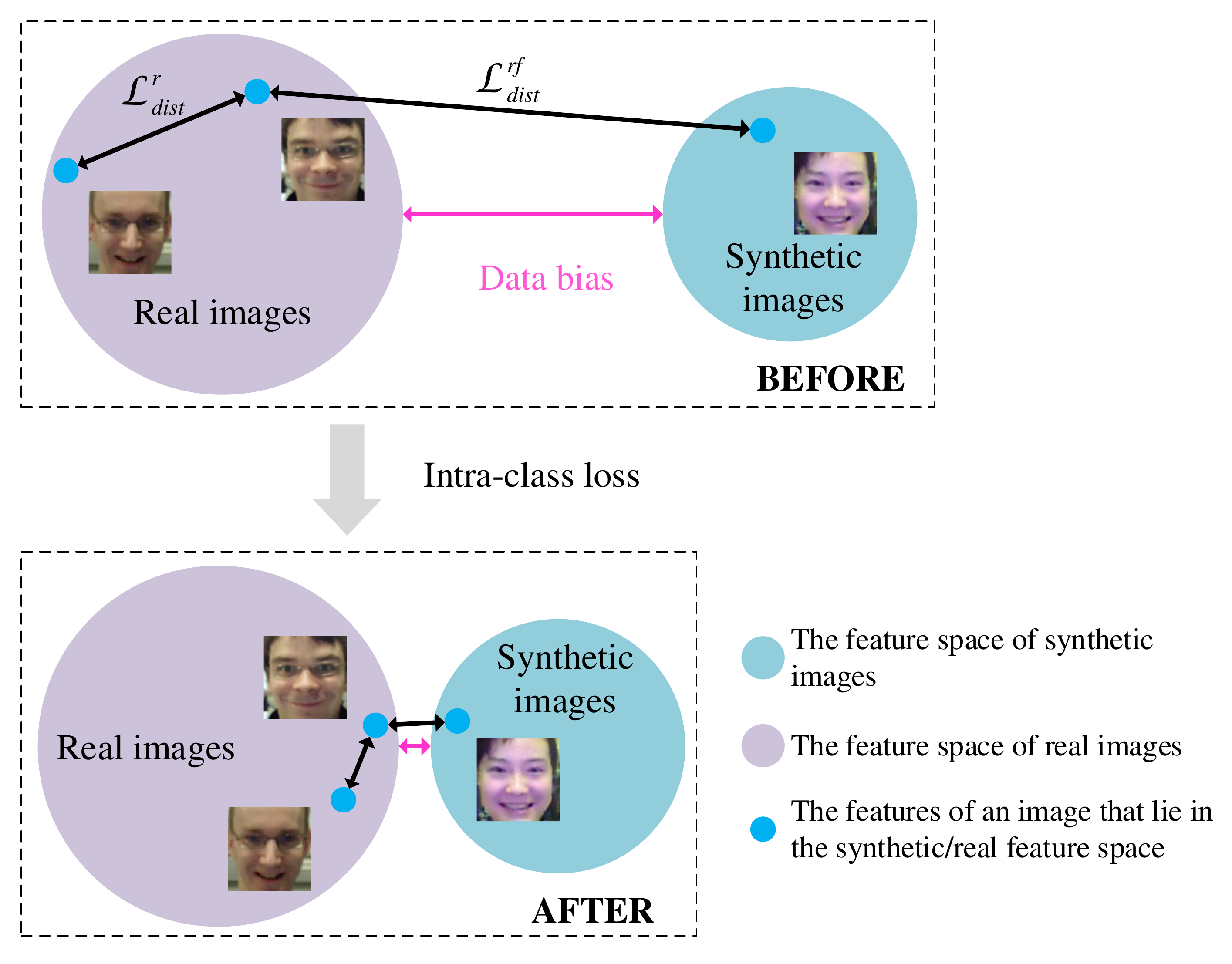}
\caption{\small Schematic illustration of the intra-class loss. The intra-class loss effectively reduces the intra-class variations of images from the same class, thus alleviating the problem of data bias between the real images and the synthetic ones.}
\label{fig:intra_loss}
\end{figure}

The intra-class loss depicts the intra-class variations of images belonging to the same class, which includes two terms, i.e., $\mathcal{L}_{dist}^{r}(\bm{x},\bm{x}^{p,r})$ between two real facial images and $\mathcal{L}_{dist}^{rf}(\bm{x},\bm{x}^{p,f})$ between a real facial image and a synthetic facial image. As shown in Fig. \ref{fig:intra_loss}, by optimizing the intra-class loss, the intra-class variations of images are minimized, and thus the problem of data bias between the real images and the synthetic images is effectively alleviated.

The traditional stochastic gradient descent (SGD) \cite{lecun1998gradient} method can be directly used to optimize the intra-class loss, which uses all gradients derived from both real and synthetic images. However, the recognition capability for the real images may be affected due to the interference of the synthetic images. Therefore, in this paper, we propose a real data-guided back-propagation (RDBP) algorithm to optimize $R_{ext}(.)$, which computes the gradients of $\mathcal{L}_{dist}^{r}$ and $\mathcal{L}_{dist}^{rf}$ in the different ways, given as follows,

\begin{equation}
\begin{split}
\label{eq:gradient_r}
\frac{\partial \mathcal{L}_{dist}^{r}(\bm{x},\bm{x}^{p,r})}{\partial W_{ext}}=&\frac{\partial  \mathcal{L}_{dist}^{r}(\bm{x},\bm{x}^{p,r})}{\partial R_{ext}(\bm{x})}\cdot \frac{\partial R_{ext}(\bm{x})}{\partial W_{ext}} \\ &+ \frac{\partial  \mathcal{L}_{dist}^{r}(\bm{x},\bm{x}^{p,r})}{\partial R_{ext}(\bm{x}^{p,r})}\cdot \frac{\partial R_{ext}(\bm{x}^{p,r})}{\partial W_{ext}},
\end{split}
\end{equation}
\begin{equation}
\label{eq:gradient_rf}
\frac{\partial \mathcal{L}_{dist}^{rf}(\bm{x},\bm{x}^{p,f})}{\partial W_{ext}}=\frac{\partial  \mathcal{L}_{dist}^{rf}(\bm{x},\bm{x}^{p,f})}{\partial R_{ext}(\bm{x}^{p,f})}\cdot \frac{\partial R_{ext}(\bm{x}^{p,f})}{\partial W_{ext}},
\end{equation}
where $W_{ext}$ is the network parameters of $R_{ext}(.)$.

The gradients derived from $\mathcal{L}_{dist}^{r}$ (Eq. (\ref{eq:gradient_r})) are normally computed, while the gradients derived from $\mathcal{L}_{dist}^{rf}(\bm{x},\bm{x}^{p,f})$ (Eq. (\ref{eq:gradient_rf})) only involve $R_{ext}(\bm{x}^{p,f})$ (the gradient $\frac{\partial  \mathcal{L}_{dist}^{rf}(\bm{x},\bm{x}^{p,f})}{\partial R_{ext}(\bm{x})}\cdot \frac{\partial R_{ext}(\bm{x})}{\partial W_{ext}}$ used in SGD is removed). Hence, in Eq. (16), $R_{ext}(\bm{x}^{p,f})$ is optimized to approach $R_{ext}(\bm{x})$ (considered as a fixed value).
In this way, the feature distribution of the synthetic images is forced to get close to that of the real images. Therefore, this effectively avoids the degeneration of the final FER performance due to interference from the synthetic facial images.

\textbf{The classification loss.} The recognition network $R$ should be capable of performing robust FER, which is our ultimate goal. Therefore, except for the intra-class loss defined above, $R$ is also trained under the supervision of the classification loss, which is defined as,
\begin{equation}
\label{eq:R_cls}
\mathcal{L}_{cls,R}=\mathcal{L}_{cls,R}^{r} + \mathcal{L}_{cls,R}^{f},
\end{equation}
where ${L}_{cls,R}^{r}$ refers to the classification loss computed for the real image $\bm{x}$, and ${L}_{cls,R}^{f}$ refers to the classification loss computed for the synthetic image $\bm{x}^{p,f}$, that is,
\begin{equation}
\mathcal{L}_{cls,R}^{r} = -\log R_{cls}(\bm{y}|R_{ext}(\bm{x})),
\end{equation}
\begin{equation}
\mathcal{L}_{cls,R}^{f} = -\log R_{cls}(\bm{y}|R_{ext}(\bm{x}^{p,f})),
\end{equation}
where $R_{cls}(\bm{y}|R_{ext}(\bm{x}))$ and $R_{cls}(\bm{y}|R_{ext}(\bm{x}^{p,f}))$ refer to the probability of correct prediction for $\bm{x}$ and $\bm{x}^{p,f}$, respectively.

\subsection{The overall learning algorithm}

The whole learning algorithm of the proposed method is composed of a two-stage learning procedure (see Algorithm \ref{alg:learning}).

In the first stage, FESGAN is pre-trained to be capable of synthesizing high-quality facial expression images. To achieve this goal, the three networks of FESGAN (i.e., $G$, $D_{img}$ and $D_{z}$), are trained with three different objective functions, which are respectively written as,
\begin{equation}
\begin{split}
\label{eq:stage1_lg}
{\mathcal{L}_{G}} {=} &{\mathcal{L}_{adv_{g}}^{img}+ \lambda_{1}\mathcal{L}_{adv_{g}}^{z} + \lambda_{2}\mathcal{L}_{rec} + \lambda_{3}\mathcal{L}_{id}} \\
&{+  \lambda_{4}\mathcal{L}_{cls,D}^{f},}
\end{split}
\end{equation}
\begin{equation}
\label{eq:stage1_ldimg}
{\mathcal{L}_{D_{img}} = \lambda_{5}\mathcal{L}_{adv_{d}}^{img} + \mathcal{L}_{cls,D}^{r},}
\end{equation}
\begin{equation}
\label{eq:stage1_ldz}
{\mathcal{L}_{D_{z}} = \mathcal{L}_{adv_{d}}^{z},}
\end{equation}
where $\lambda_{1}$, $\lambda_{2}$, $\lambda_{3}$, $\lambda_{4}$ are the  hyper-parameters for balancing $\mathcal{L}_{G}$. $\lambda_{5}$ is the hyper-parameter for balancing $\mathcal{L}_{D_{img}}$.

In the second stage, the pre-trained FESGAN and the recognition network $R$ are jointly trained, where both the real and synthetic facial images are used for training. To make the synthetic images generated by FESGAN be beneficial to the recognition network $R$, the classification loss $\mathcal{L}_{cls,R}^{f}$ computed by $R$ for the synthetic images is used to supervise the training of $G$ in FESGAN. Therefore, the objective function for $G$ is rewritten as,
\begin{equation}
\label{eq:stage2_lg}
\begin{split}
{\mathcal{L}_{G} = }&{\mathcal{L}_{adv_{g}}^{img} + \lambda_{1}\mathcal{L}_{adv_{g}}^{z} + \lambda_{2}\mathcal{L}_{rec} + \lambda_{3}\mathcal{L}_{id}} \\
&{+ \lambda_{4}(\mathcal{L}_{cls,D}^{f} + \mathcal{L}_{cls,R}^{f})}
\end{split}
\end{equation}
and the objective function for optimizing $R$ is written as
\begin{equation}
\label{eq:stage2_lr}
\mathcal{L}_{R} = \lambda_{6}\mathcal{L}_{intra}+\mathcal{L}_{cls,R},
\end{equation}
where $\lambda_{6}$ is the hyper-parameter for balancing $\mathcal{L}_{R}$.

In general, four networks (i.e., $G$, $D_{img}$, $D_{z}$ and $R$) are jointly trained with different objective functions in the second stage.

\begin{algorithm}[t]
\small
\caption{\small The learning algorithm of the proposed method.}
\hspace*{0in}{\bf Input:}
Training data $\{\bm{x},\bm{x}^{p,r},\bm{y},\bm{y'}\}$. Initialized parameters $W_{G}$, $W_{D_{img}}$, $W_{D_{z}}$ and $W_{R}$ for $G$, $D_{img}$, $D_{z}$ and $R$, respectively. The number of iterations for pre-training FESGAN $P_{pre}$. The maximum number of iterations $P_{max}$. \\
\hspace*{0in}{\bf Output:}
The estimated $\hat{W}_{G}$, $\hat{W}_{D_{img}}$, $\hat{W}_{D_{z}}$ and $\hat{W}_{R}$.
\begin{algorithmic}[1]
\State $t=0$;
\While{$t<P_{max}$}
	\State Compute the face representation $g(\bm{x})$ according to Eq. (\ref{eq:gx});
	\State Randomly sample a vector $\bm{z}$ from $U(-1,1)$;
	\State Compute $\mathcal{L}_{D_{z}}$ according to Eq. (\ref{eq:stage1_ldz});
	\State Update $W_{D_{z}}$ for $D_{z}$;
	\State Obtain the synthetic image $\bm{\hat{x}}$ according to Eq. (\ref{eq:xb});
	\State Compute $\mathcal{L}_{D_{img}}$ according to Eq. (\ref{eq:stage1_ldimg});
	\State Update $W_{D_{img}}$ for $D_{img}$;
	\State Obtain the reconstructed image $\bm{x}^{rec}$ according to Eq. (\ref{eq:xrec});
	\If{$t<P_{pre}$}
\State Compute $\mathcal{L}_{G}$ according to Eq. (\ref{eq:stage1_lg});
	\Else
			\State Obtain the synthetic image $\bm{x}^{p,f}$ generated from $\bm{z}$ and $\bm{y}$ according to Eq. (\ref{eq:xpf});
		\State Compute $\mathcal{L}_{R}$ according to Eq. (\ref{eq:stage2_lr});
		\State Update $W_{R}$ for $R$;
		\State Compute $\mathcal{L}_{G}$ according to Eq. (\ref{eq:stage2_lg});
	\EndIf
	\State Update $W_{G}$ for $G$;
	\State $t=t+1$;
\EndWhile
\end{algorithmic}
\label{alg:learning}
\end{algorithm}

\begin{table*}
\caption{\small Architecture of FESGAN, where Conv($n,m,s$) and DeConv($n,m,s$) respectively denote the convolutional layer and the transposed convolutional layer with the number of output feature maps $n$, kernel size $m\times m$ and stride $s$. FC($n$) refers to the fully-connected layer with the output features of $n$ dimensions. LReLU represents the leaky ReLU \cite{maas2013rectifier}. IN represents the instance normalization \cite{ulyanov2016instance}. The values of $c$ and $K$ refer to the channels of the input image and the number of classes, respectively.}
\label{tab:fesnet}
\centering
\begin{tabular}{|c|c|c|c|c|}
\hline
$G_{enc}$ & $G_{dec}$ & \multicolumn{2}{c|}{$D_{img}$} & $D_{z}$ \\
\hline
Conv(64,4,2), IN, LReLU & DeConv(512,4,2), IN, ReLU & \multicolumn{2}{c|}{Conv(64,4,2), LReLU} & FC(64), LReLU \\
\hline
Conv(128,4,2), IN, LReLU & DeConv(256,4,2), IN, ReLU & \multicolumn{2}{c|}{Conv(128,4,2), LReLU} & FC(32), LReLU \\
\hline
Conv(256,4,2), IN, LReLU & DeConv(128,4,2), IN, ReLU & \multicolumn{2}{c|}{Conv(256,4,2), LReLU} & FC(16), LReLU \\
\hline
Conv(512,4,2), IN, LReLU & DeConv(64,4,2), IN, ReLU & \multicolumn{2}{c|}{Conv(512,4,2), LReLU} & FC(1), Sigmoid \\
\hline
Conv(1024,4,2), IN, LReLU & DeConv(32,4,2), IN, ReLU & Conv(1,4,2) ~($D_{img}^{adv}$)& Conv(1024,4,2), LReLU & \\
\hline
FC(64), Tanh & DeConv($c$,4,2), Tanh &  & Conv($K$,4,1)~($D_{cls}^{adv}$)& \\
\hline
\end{tabular}
\end{table*}

\subsection{Discussions}
It is worth noting that using the classification loss from the recognition network may reduce the diversity and recognition difficulty of the generated facial images, if the tasks of facial expression synthesis and facial expression recognition are not properly combined and learned.

In this paper, the proposed method can effectively deal with the above issue from two aspects.
Firstly, FESGAN is designed to generate new training samples with new facial identities. To achieve this, the discriminator $D_{img}$ is adopted to discriminate the latent face representation $g(\bm{x})$ and the random vector $\bm{z}$ sampled from the prior distribution. As a result, the trained FESGAN is able to generate new samples (with new facial identities) using the random vectors. Such a way significantly increases the diversity of the generated facial images. Therefore, the training data can be effectively augmented based on the limited training samples.
Secondly,
we leverage a two-stage learning procedure for the joint learning of FSEGAN and the recognition network $R$.
In the first stage,  FESGAN is pre-trained to be capable of synthesizing high-quality facial expression images, where the generator considers the classification loss from $D_{img}$ (note that $D_{img}$ is employed in the FESGAN and is independent of the recognition network $R$). Therefore, the generator is able to generate facial expression images without the influence of the recognition network.
  In the second stage, the pre-trained FESGAN and $R$ are jointly optimized, where the generator employs the classification losses from $R$ and $D_{img}$.  In other words, the classification loss obtained from the recognition network is only used in the second stage of joint learning, which makes the generated facial images beneficial for classification. The second stage plays the role similar to `fine-tuning', which aims to boost the performance of two tasks. Thus, the generator will not be greatly affected by the recognition network, since it needs to balance the losses from $R$, $D_z$ and $D_{img}$.
\section{Experiments}

In this section, we firstly introduce three popular facial expression databases in Section \ref{sec:database} and the implementation details in Section \ref{sec:implementation}. Then, we discuss the effectiveness of FESGAN for facial expression synthesis in Section \ref{sec:fes}. We give an ablation study of the proposed method for FER in Section \ref{sec:abl_study}. Finally, we compare the performance of the proposed method with several state-of-the-art FER methods in Section \ref{sec:cmp_state} and Section \ref{otherdatasets}.

\subsection{Databases and Protocols}
\label{sec:database}

To show the effectiveness of our proposed method, we conduct extensive experiments on five popular databases, including CK+ \cite{lucey2010extended}, Oulu-CASIA \cite{zhao2011facial}, MMI \cite{valstar2010induced}, Multi-PIE \cite{Gross2010} and TFD  \cite{Susskind2010}.

\textbf{CK+ database} The CK+ database \cite{lucey2010extended} consists of 593 image sequences with 123 subjects, among which 327 image sequences containing 118 subjects are labeled with seven facial expressions (i.e., anger, contempt, disgust, fear, happiness, sadness and surprise). Each sequence starts with the neutral facial expression to the apex of a specified facial expression and only the last frame of the sequence is provided with the label. In this paper, we employ the popular ten-fold cross validation protocol as \cite{jung2015joint,zhang2017facial} for evaluation.

\textbf{Oulu-CASIA database} The Oulu-CASIA database \cite{zhao2011facial} consists of 480 image sequences with six facial expressions (i.e., anger, disgust, fear, happiness, sadness and surprise), which contains 80 subjects with the age from 23 to 58. Similar to the CK+ database, all the sequences begin from the neutral facial expression and end with the peak facial expression. In our experiments, we adopt the ten-fold cross validation protocol in the subject-independent way as the CK+ database.

\textbf{MMI database} The MMI database \cite{valstar2010induced} consists of 205 image sequences, which are captured in the frontal view and labelled with six facial expressions (i.e., anger, disgust, fear, happiness, sadness and surprise). There are 30 subjects aged from 19 to 62. Different from CK+ and Oulu-CASIA, each sequence in the MMI database starts with a neutral facial expression and ends with a neutral facial expression, where the peak facial expression appears  in the middle of each sequence. Since the MMI database does not label the frame for the peak facial expression, we manually select the frame showing the peak facial expression. This database is very challenging, where the facial expressions are non-uniformly posed and some accessories (e.g., glasses, caps) are non-linearly coupled in many faces. We also conduct the subject-independent ten-fold cross validation as CK+ for this database.

\textbf{Multi-PIE database} The Multi-PIE database \cite{Gross2010} contains a large number of facial images captured under various pose and illumination conditions along with different facial expressions. Following the same experimental settings as \cite{Eleftheriadis2015}, we use the images of 270 subjects with six facial expressions (i.e., neutral, disgust, surprise, smile, scream and squint) and five pose variations (including $-30^{\circ}$, $-15^{\circ}$, $0^{\circ}$, $15^{\circ}$ and $30^{\circ}$). For each pose, we have 1531 images. Therefore, we have $1531 \times5 = 7655$ facial images in total. We conduct the subject-independent five-fold cross validation for this database as in \cite{Eleftheriadis2015}.

\textbf{TFD database} The TFD database \cite{Susskind2010} is a large facial expression database, which is comprised of images from several facial expression databases. It
contains 4178 images with seven facial expressions (i.e., anger, disgust, fear, happiness, sadness, surprise and neutral). The images are divided into 5 individual folds, each of which contains a training set, a validation set, and a test
set consisting of 70\%, 10\%, and 20\% of the images, respectively. We train
the proposed model using the training sets and report the average
results over five folds on the test sets as in \cite{Ding2017}.

\subsection{Implementation Details}
\label{sec:implementation}

\begin{table}
\caption{\small Architecture of the recognition network $R$}
\label{tab:rnet}
\centering
\begin{tabular}{|c|c|}
\hline
\multirow{4}{*}{$R_{ext}$} & Conv1-Pool3 from Light CNN-29, \textbf{frozen}\\
\cline{2-2}
 & Conv(256,3,2), ReLU\\
\cline{2-2}
 & Conv(512,3,2), ReLU\\
\cline{2-2}
 & FC(2048), ReLU\\
\cline{2-2}
 & FC(512), ReLU, Dropout(0.5) \\
\hline
$R_{cls}$ & FC($K$) \\
\hline
\end{tabular}
\end{table}

Table \ref{tab:fesnet} shows the detailed architecture of each network in FESGAN. The encoder $G_{enc}$ is composed of five convolutional layers and a fully-connected layer to map the input into a vector (i.e., the face representation). The decoder $G_{dec}$  generates the synthetic image with $c$ channels from the input face representation and a label vector via six transposed convolutional layers. We use the instance normalization for both $G_{enc}$ and $G_{dec}$. The discriminator $D_{img}$ firstly processes the input using four convolutional layers, and then generates the outputs using a convolutional layer (corresponding to adversarial learning) and two convolutional layers (corresponding to classification).
The discriminator $D_{z}$ uses four fully-connected layers to determine whether the input is sampled from the uniform distribution. The adversarial losses between $G$ and $D_{img}$ (i.e., $\mathcal{L}_{adv_{d}}^{img}$ and $\mathcal{L}_{adv_{g}}^{img}$) are optimized by using WGAN-GP \cite{gulrajani2017improved}.
Table \ref{tab:rnet} shows the detailed architecture of the recognition network $R$, where $R_{ext}$ is composed of a part of the Light CNN-29 \cite{wu2018light}, two convolutional layers and two fully-connected layers to extract features, and $R_{cls}$ is a fully-connected layer to perform FER. 
Note that, the weights of the layers from the Light CNN-29 are frozen all the time during the training process.

For all the databases, we align each face using the landmarks detected by Dlib\footnote{http://dlib.net}, and then crop and resize the face into the size of $128\times 128$.
To collect more data, we use the last three frames of the sequences for the CK+ and Oulu-CASIA databases
(for the MMI database, we manually choose three adjacent frames at the apex intensity)
for training and use the peak frame for testing. We train the networks using the Adam algorithm \cite{kingma2015adam} with the learning rate of 0.0001, $\beta_{1}=0.5$ and $\beta_{2}=0.999$. The coefficients $\lambda_{1}$, $\lambda_{2}$, $\lambda_{3}$, $\lambda_{4}$, $\lambda_{5}$, and $\lambda_{6}$ are empirically set to 1, 10, 5, 1, 1 and 0.001, respectively. These parameters are kept fixed for all the experiments.
The proposed method is implemented based on Pytorch\footnote{http://pytorch.org} and all the models are trained on a single NVIDIA GTX TITAN GPU with the batch size of 16.

\begin{figure}[!t]
\centering
\includegraphics[width=3.65in]{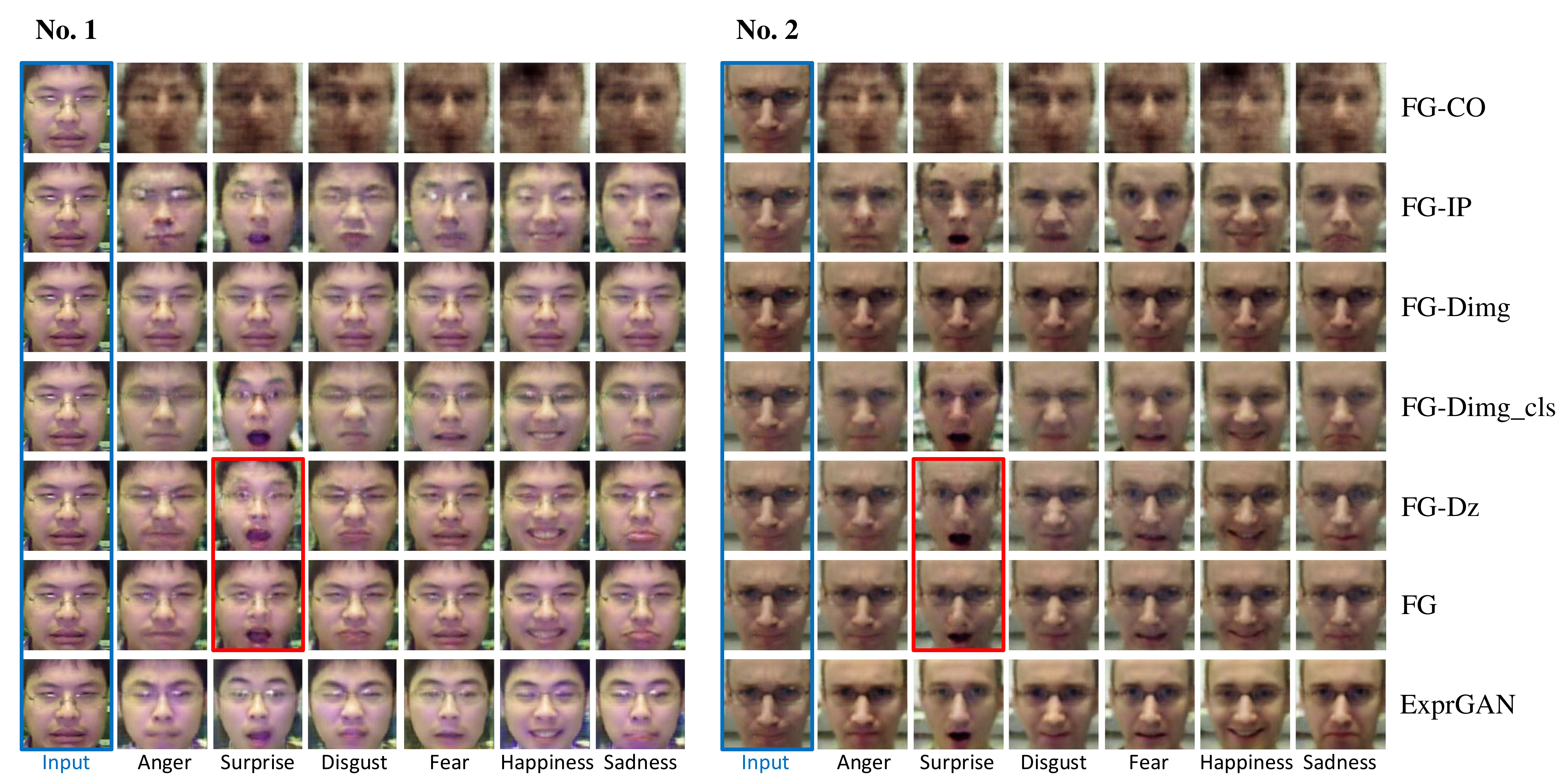}
\caption{\small Results of facial expression synthesis obtained by different variants of the proposed FESGAN on the Oulu-CASIA database.}
\label{fig:fesgan_abla}
\end{figure}

\subsection{Facial Expression Synthesis}
\label{sec:fes}

In this subsection, we demonstrate the capability of FESGAN to synthesize facial expression images. The synthetic results are shown from Figs. \ref{fig:fesgan_abla} to 7. The CK+, Oulu-CASIA, MMI databases are employed.

\subsubsection{The effectiveness of different components} Fig.~\ref{fig:fesgan_abla} shows the synthetic results obtained by different variants of the proposed FESGAN. Specifically, FG denotes the proposed FESGAN as we describe in Sec \ref{sec:fesgan}. FG-CO denotes the proposed FESGAN trained without using content learning (i.e., the reconstruction loss $\mathcal{L}_{rec}$ and the identity preserving loss $\mathcal{L}_{id}$ are not used). FG-IP denotes the proposed FESGAN trained without using the identity preserving loss $\mathcal{L}_{id}$. FG-Dimg denotes the proposed FESGAN trained without using the discriminator $D_{img}$. FG-Dimg\_cls denotes the proposed FESGAN trained without using the auxiliary classifier in $D_{img}$. FG-Dz denotes the proposed FESGAN trained without using the discriminator $D_{z}$. Besides, we also compare the recently-proposed ExprGAN \cite{ding2018exprgan}, which can also effectively generate different facial expression images based on GAN.

From Fig.~\ref{fig:fesgan_abla}, we have the following observations. The synthetic images generated by FG-CO show the worst image quality among all the variants, where the appearance information of the input images is not effectively learned. The results obtained by FG-IP have much better image quality than those obtained by FG-CO. This is because that the reconstruction loss is helpful to preserve the content of the input image. However, the identities in these images (obtained by FG-IP) are changed, which demonstrates the importance of the identity preserving loss. The generated images by FG-Dimg show the same facial expression as the input image. This is due to the fact that FESGAN degenerates to an auto-encoder (AE) without using the discriminator $D_{img}$.

The results obtained by FG-Dimg\_cls show that, the generator can still synthesize the facial images with different expressions when the  auxiliary classifier in $D_{img}$ is not used.
However, compared with the input image, some details in these synthetic images may be lost. For example, the synthetic images show blurred textures with the changed color style for the No.~1 person. The glasses in the synthetic images are removed for the No.~2 person.
The comparison between FG-Dimg\_cls and FG shows that the auxiliary classifier plays a critical role in learning the facial expression and facial identity for the generator. This is mainly because that the auxiliary classifier in $D_{img}$ makes the generator disentangle the expression information from the other information and learn the relationship between the expression information and the input label during synthesizing facial expression images.

Compared with the other variants, FG, FG-Dz and ExprGAN show good performance of facial expression synthesis.
Nevertheless, from the results obtained by FG-Dz, the variations between the synthetic images of different expressions may not be smooth, which affects the details of the facial image. For example, the synthetic images with the surprise expression generated by FG-Dz for both the No.~1 and No.~2 persons show unexpected distortion (the facial textures are damaged for the No.~1 person, while the glasses are removed for the No.~2 person). In contrast, the synthetic images generated by FG give more smooth transformation than those generated by FG-Dz, which demonstrates that the discriminator $D_{z}$ is beneficial to compact the face representations on the facial manifold \cite{zhang2017age}. Compared with FG, the results obtained by ExprGAN sometimes cannot effectively preserve identity information and facial details.
In addition, as we demonstrate in the following section (see Fig.~7), the discriminator $D_{z}$ plays a critical role in generating images with new identities from a prior distribution.

\begin{figure}[!t]
\centering
\includegraphics[width=2.5in]{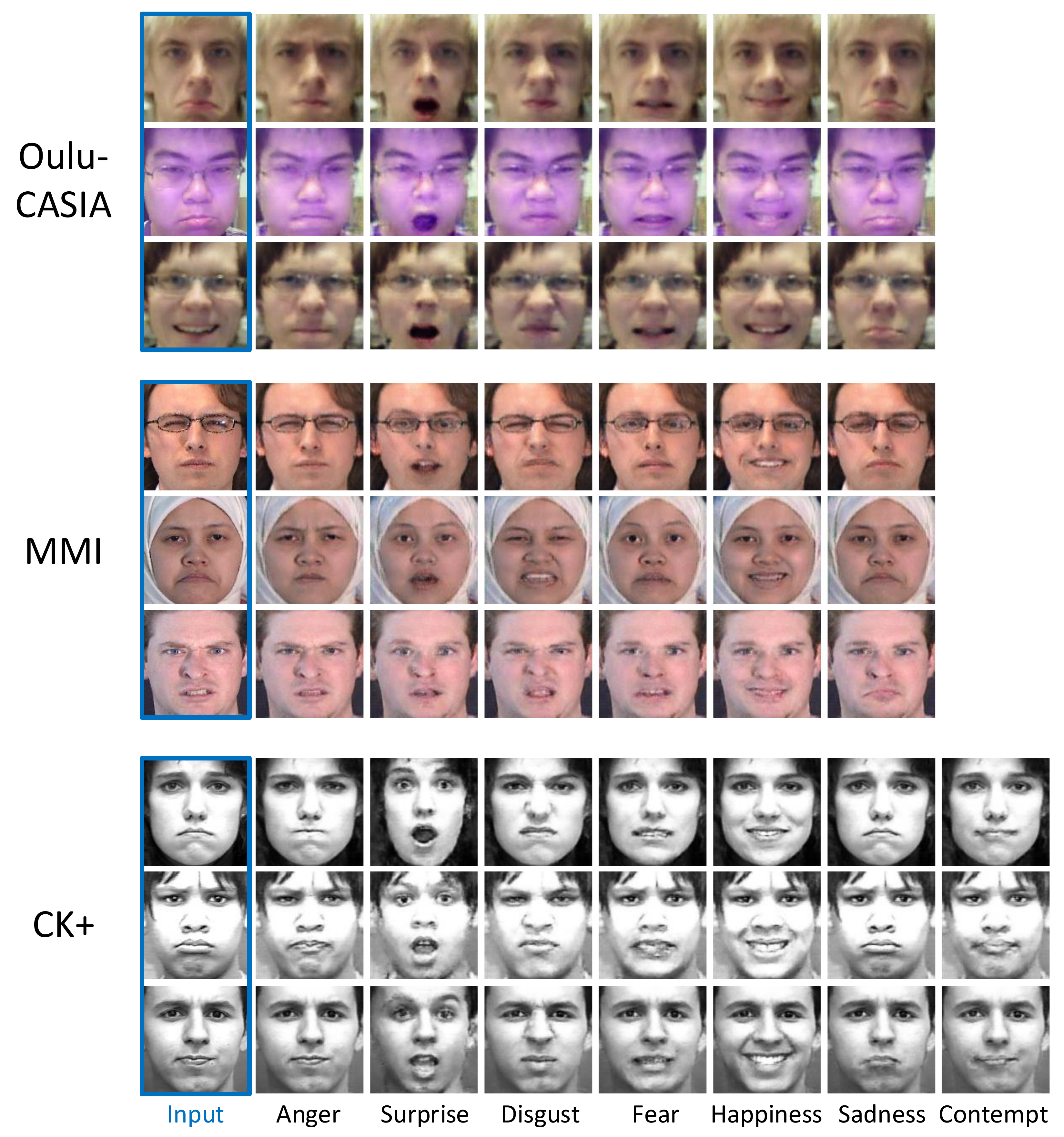}
\caption{\small More results of facial expression synthesis. From left to right, the input images, the generated images with different expression categories.}
\label{fig:all_transfer}
\end{figure}

Fig. \ref{fig:all_transfer} shows more synthetic results for different facial expressions on three databases. We can observe that FESGAN is able to effectively change the categories of facial expression for the input images on these three databases, while sufficiently preserving most of facial expression-unrelated contents.

We also quantitatively compare the generation performance between our proposed  FESGAN and two state-of-the-art facial expression synthesis methods (including G2-GAN and ExprGAN). Two objective
image quality metrics (i.e., the peak-signal-to-noise ratio (PSNR) and the structural similarity index measure (SSIM)) are employed for performance comparison.
Specifically, for each real image in the test set, we firstly generate several synthetic images corresponding to different expressions.  Then, PSNR (or SSIM) is computed between each synthetic image and the ground-truth image in the test set (both images have the same facial expression and the same facial identity). Finally, the average PSNR (or SSIM) is obtained as the final result.
Table \ref{tab:psnr} shows the comparison results. Note that the PSNR and SSIM obtained by G2-GAN are taken from \cite{qiao2018geometry}.

\begin{table}[!t]
\caption{\small Quantitative results for facial expression synthesis on the CK+, Oulu-CASIA and MMI databases. ``-'' indicates that the corresponding results are not provided by the method. The best results are boldfaced.}
\label{tab:psnr}
\centering
\scalebox{1.2}{
\begin{tabular}{|c|c|c|c|}
\cline{1-4}
Databases &Methods & PSNR & SSIM \\
\hline

\multirow{3}{*}{CK+} & ExprGAN & 19.654 & 0.575\\
\cline{2-4} & G2-GAN & \textbf{24.420} & \textbf{0.767} \\
\cline{2-4} & FESGAN & 21.380 & 0.698\\
\hline
\multirow{3}{*}{Oulu-CASIA} & ExprGAN & 19.982 & 0.648 \\
\cline{2-4} & G2-GAN & \textbf{26.588} & \textbf{0.914} \\
\cline{2-4} &  FESGAN & 22.079 & 0.754\\
\hline
\multirow{3}{*}{MMI} & ExprGAN & 21.012 & 0.565 \\
\cline{2-4} & G2-GAN & -- & -- \\
\cline{2-4} &  FESGAN &\textbf{24.676} & \textbf{0.761} \\
\hline
\end{tabular}}
\end{table}

From Table \ref{tab:psnr}, we can see that FESGAN obtains worse performance than G2-GAN in terms of both PSNR and SSIM.
This is mainly because that the commonly-used skip connections (which are beneficial to generate high-quality facial images having the similar underlying structure to the input) between the encoder and the decoder \cite{song2018geometry,qiao2018geometry} are not adopted in our method.  In this paper, FESGAN is designed to generate facial images with new identities, so that the diversity of training data can be effectively enhanced. To achieve this, the random vector sampled from the uniform distribution is used as the input of $G_{dec}$ (the decoder of the generator in FESGAN).
As a result, the skip connections cannot be used in our method, since they require the images as the input. Therefore, the details of the
generated facial images may not be preserved as well as those
in \cite{song2018geometry,qiao2018geometry}.
Compared with ExprGAN that is also not used the skip connections, FESGAN obtains higher PSNR and SSIM. This can be ascribed to the effectiveness of the network architecture of FESGAN.

\begin{figure}[!t]
\centering
\includegraphics[width=2.5in]{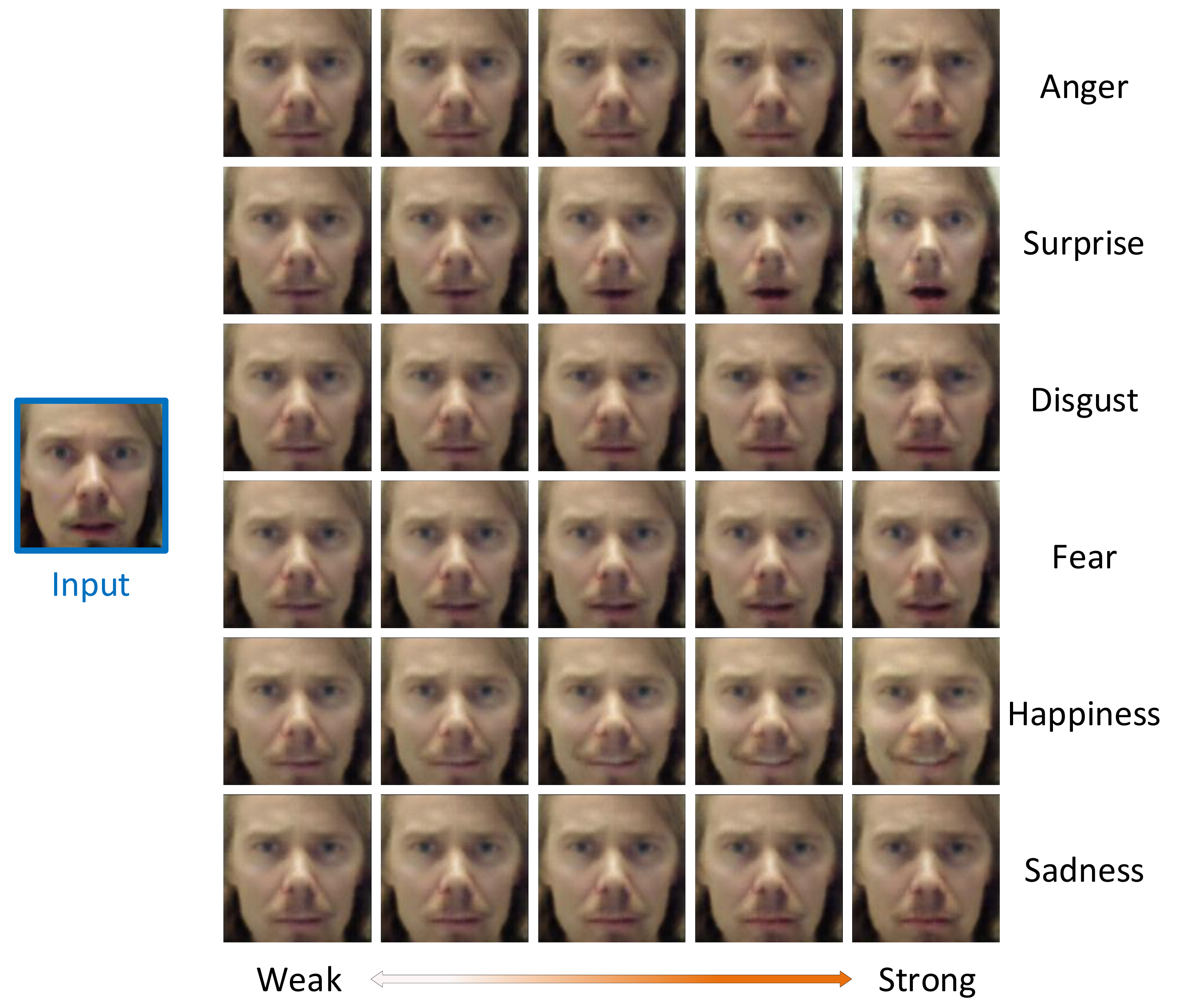}
\caption{\small Results of facial expression synthesis with different categories and intensities of facial expressions on the Oulu-CASIA database. From left to right, the input image, the synthetic images with five different intensities from weak to strong. Each row represents an expression category.}
\label{fig:oulu_inten}
\end{figure}

\subsubsection{Expression intensity modelling} Fig. \ref{fig:oulu_inten} shows the synthetic results for different categories and intensities of facial expressions on the Oulu-CASIA database, where each row represents a specified facial expression category with different intensities. This is done by sliding $u(\bm{y'})$ on $[-1,1]$ to control the intensity for the target category. As we can see, the changes between faces in each row are gradual and natural, which indicates that FESGAN can successfully model the intensity given the facial expressions.

On one hand, it is also worth pointing out that the synthetic images on the Oulu-CASIA database are not very satisfactory (some synthesized expressions, such as anger, fear and sadness, are not very distinguishable) in Fig.~4. Moreover, the change of expression from strong to weak is not significant in Fig.~6. This can be ascribed to the fact that the number of training samples is relatively limited (1,296 images in total), and the differences between some expressions (such as anger and sadness) are subtle on the Oulu-CASIA database. In addition, as we mention before, the commonly-used skip connections between the encoder and the decoder are not adopted.

On the other hand, these generated images (with new facial identities) can be effectively
used as the auxiliary training data that are beneficial to the
recognition network $R$. This is mainly because the proposed intra-class loss effectively mitigates the problem of data bias between the real facial images and the
synthetic facial images in the recognition network. Furthermore, the proposed RDBP algorithm is further leveraged to optimize the intra-class loss, avoiding the interference from the synthetic facial images. Experimental results (see Sections IV-D-4) and IV-D-5)) verify the importance of both intra-class loss and RDBP for improving the performance of expression recognition.

In a word, although the quality of synthetic images may not be perfectly satisfactory,
these synthetic images can be used to effectively boost the overall performance by taking advantage of the proposed intra-class loss and its corresponding optimization algorithm (RDBP).
\subsubsection{Generating images from a prior distribution} 
\begin{figure}[!t]
\centering
\includegraphics[width=2.2in]{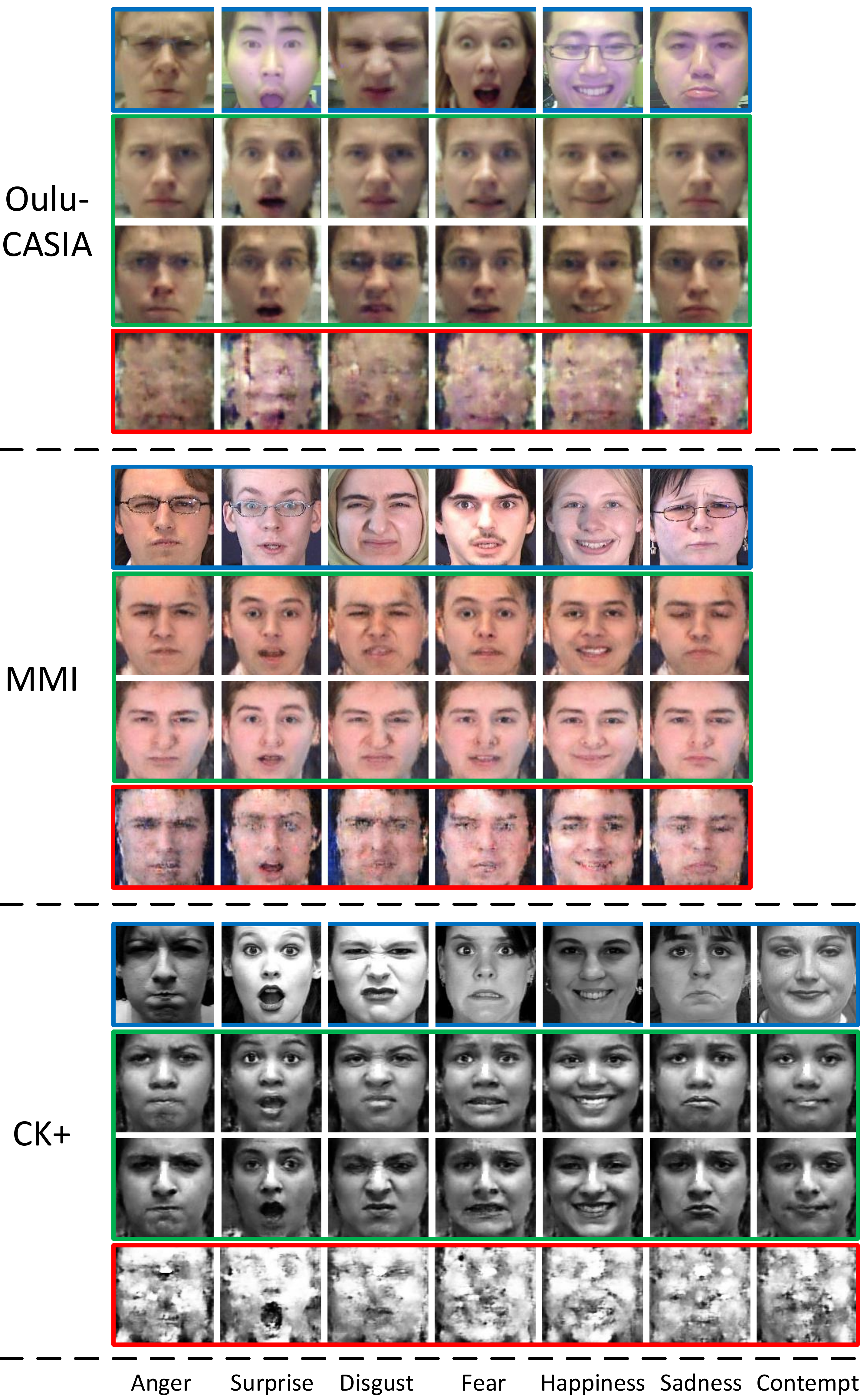}
\caption{\small Comparison between the real facial images and the synthetic images (obtained by FG and FG-Dz, respectively). These synthetic images are generated from the random vectors sampled from $U(-1,1)$. From top to bottom for each database, the real images, the synthetic images obtained by FG, the synthetic images obtained by FG-Dz.}
\label{fig:all_rand}
\end{figure}

Fig. \ref{fig:all_rand} shows the comparison between the real images and the synthetic images, where the second row and the third row in each database refer to the results generated by FG and FG-Dz, respectively. We can observe that through the adoption of $Dz$, FESGAN is capable of generating facial images with new identities from a prior distribution, and these synthetic images have the similar content style to the real images from the database. However, these synthetic faces obtained by FG-Dz show serious distortions.

\subsubsection{Face verification} The visual comparison between FG and FG-IP in Fig.~\ref{fig:fesgan_abla} shows the superiority of FG, which adopts the identity preserving loss. In this section, we quantitatively validate the effectiveness of the identity preserving loss by performing the task of face verification on the three databases.

More specific, for each database, we construct the facial image pairs, each of which contains a real image and a generated image with the same identity or not. Specifically, for a real image, six (or seven) facial images (corresponding to different facial expressions) having the same identity as the real image, and six (or seven) facial images having the different identities from the real image, are generated by using FESGAN. In total, 12 (or 14) synthetic images are generated for each real image. The Light CNN-29 is used as the extractor of facial identity and the cosine distance is employed for metric comparison.
The final verification rate is used for comparison.
 Table \ref{tab:face_verification} shows the face verification results obtained by FG (using the identity preserving loss) and FG-IP (without using the identity preserving loss).
 We can see that the verification rates obtained by FG on all databases are far higher than those obtained by FG-IP, which shows the effectiveness of the identity preserving loss.

\begin{table}[!t]
\caption{\small Verification rates (\%) obtained by FG and FG-IP on the CK+, Oulu-CASIA and MMI databases.}
\label{tab:face_verification}
\centering
\scalebox{1.2}{
\begin{tabular}{|c|c|c|}
\cline{1-3}
Databases &Methods & Verification Rate  \\
\hline

\multirow{2}{*}{CK+} & FG-IP & 55.15 \\
\cline{2-3}  &FG & \textbf{93.33}\\
\hline
\multirow{2}{*}{Oulu-CASIA} & FG-IP & 55.62 \\
\cline{2-3}  &FG & \textbf{99.41}\\
\hline
\multirow{2}{*}{MMI} & FG-IP & 60.95 \\
\cline{2-3}  &FG & \textbf{99.41} \\
\hline
\end{tabular}}
\end{table}

\subsection{Ablation Study for Facial Expression Recognition}
\label{sec:abl_study}

In this subsection, in order to show the effectiveness of the proposed method, we give an ablation study to evaluate different variants of the proposed method for FER. The CK+, Oulu-CASIA, MMI databases are employed.

Specifically, the proposed method, which performs joint deep learning of facial expression synthesis and recognition, is denoted as FESR\_JL. The recognition network $R$ trained only with the classification loss on the real images is used as the baseline (denoted as BASELINE). The proposed method, which performs facial synthesis and facial expression recognition in two separated CNNs, is denoted as FESR\_SL.
The proposed method, which jointly performs facial expression synthesis and facial expression recognition based on one-stage learning (i.e., the first stage for pre-training
FESGAN is not used), is denoted as FESR\_OneSt.
The proposed method, which does not use the proposed intra-class loss, is denoted as FESR\_JL-IL. Finally, the proposed method, which uses the standard SGD (instead of the proposed RDBP), is denoted as FESR\_JL-RDBP.

\begin{figure}[!t]
\centering
\includegraphics[width=2.8in]{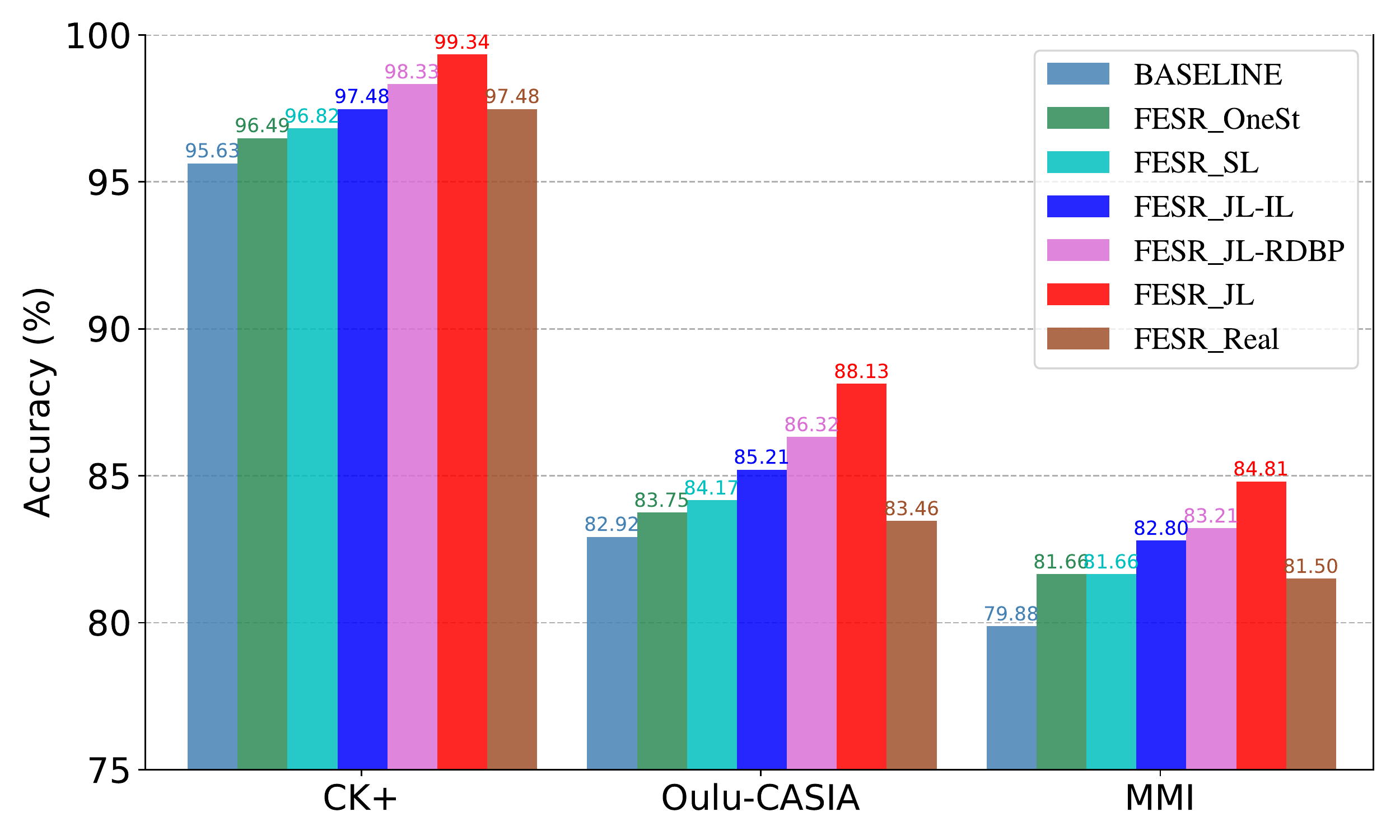}
\caption{\small Recognition results obtained by different variants of the proposed method.}
\label{fig:fesr_cmp}
\end{figure}

In addition, to validate that the synthetic images with new identities (generated from a prior distribution) can effectively enhance the performance of $R$, we further evaluate another variant of the propose method, where the synthetic images generating from $g(\bm{x})$ instead of $\bm{z}$ are used as the auxiliary training data for $R$ (denoted as FESR\_Real). In particular, we generate images with all expressions from the input images during the joint learning of facial expression synthesis and recognition. These synthetic images are combined with the original input images as the training data. Note that the only difference between FESR\_Real and FESR\_JL-IL is the strategy of  generating the auxiliary training data.

Fig. \ref{fig:fesr_cmp} shows the performance comparison among all the variants. We can see that the BASELINE performs worst on all the databases. Therefore, using the synthetic images as the auxiliary training data can effectively improve the final performance. In the following subsections, we will respectively discuss the effectiveness of the joint learning, the intra-class loss, the RDBP algorithm and the synthetic images with new identities.

\subsubsection{The effectiveness of the joint learning.}

As shown in Fig. \ref{fig:fesr_cmp}, we can see that all the variants based on the joint learning achieve better performance than FESR\_SL, which shows that the joint learning is beneficial to improve the performance of FER.
Although the synthetic images generated by FESGAN are visually photo-realistic in FESR\_SL, they may not be useful for the final recognition.
This is mainly because that the considerable redundant information implicitly exists in the generated images, which leads to the data bias between the real images and the synthetic images.
In contrast, during the joint learning of FESGAN and $R$ in FESR\_JL, the classification loss computed by $R$ for the synthetic images is used to not only optimize $R$, but also conversely supervise the optimization of $G$ in FESGAN. Such a way effectively encourages $G$ to adjust itself to generate facial images that are beneficial to the training of $R$. By the joint learning, the redundant information in the synthetic images is effectively reduced. Therefore, $R$ can make better use of the valuable information to improve the performance of FER.

\subsubsection{The effectiveness of the intra-class loss}
\label{sec:effect_il}

The performance obtained by FESR\_JL is better than that obtained by FESR\_JL-IL, which does not consider the intra-class loss. On one hand, the joint learning of facial expression synthesis and recognition can be viewed as an effective strategy of alleviating the mismatch issue between FESGAN and the recognition network $R$ (i.e., the separate training of CNNs will degrade the performance as shown in Section IV-D-1). On the other hand, the intra-class is designed to mitigate the problem of data bias between the real images and the synthetic images, which plays an important role in improving the performance of the recognition network $R$.

In this paper, the intra-class loss takes three images from the same class as  the input, which involves two real images and one synthetic image. By reducing the intra-class variations of these images from the same class, the distributions of the corresponding features are effectively compacted, and thus the problem of data bias between the real images and the synthetic images is significantly alleviated.

\begin{figure}[!t]
\centering
\subfloat[]{\includegraphics[width=1.7in]{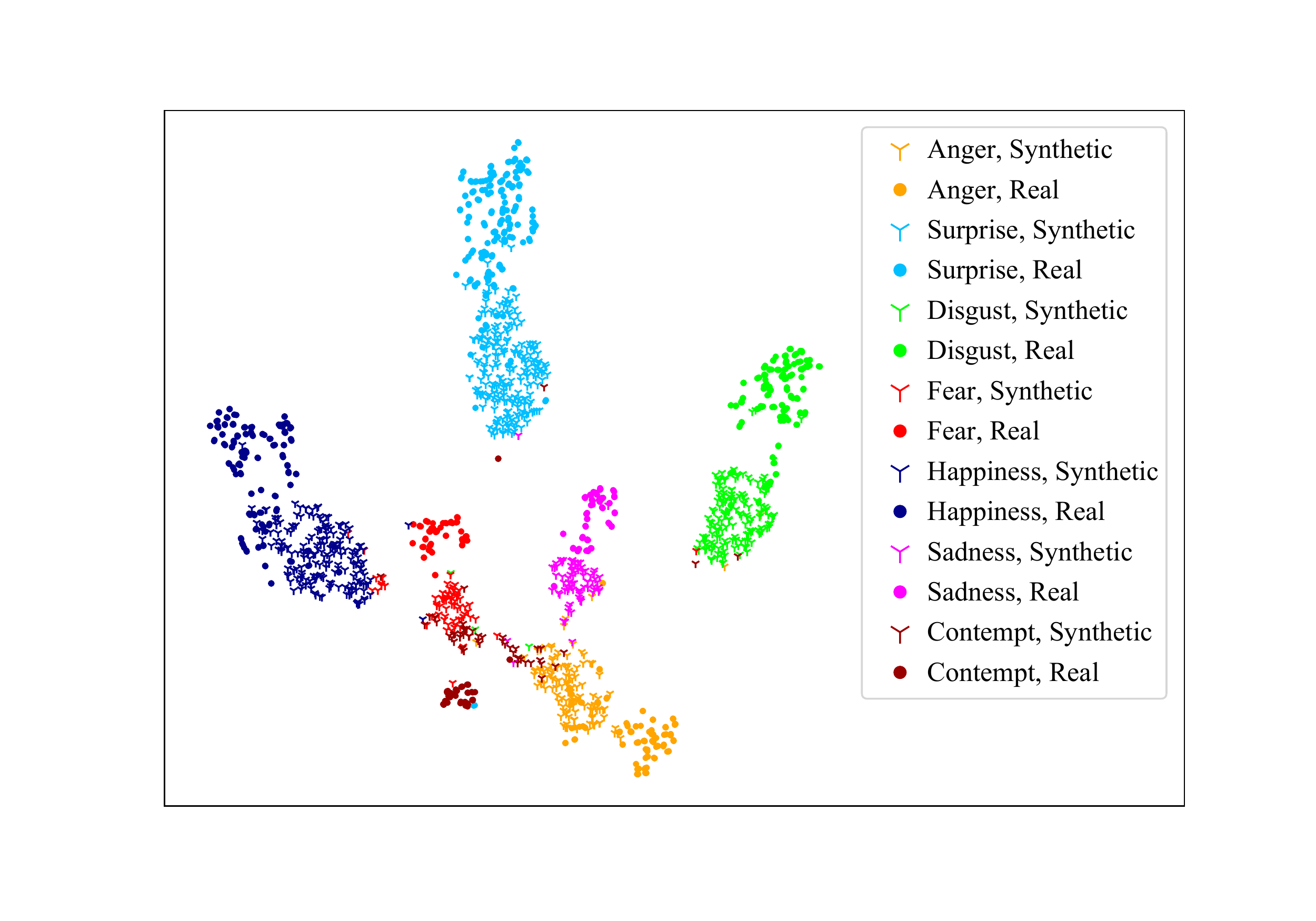}%
\label{fig:without_IL_allexp}}
\subfloat[]{\includegraphics[width=1.7in]{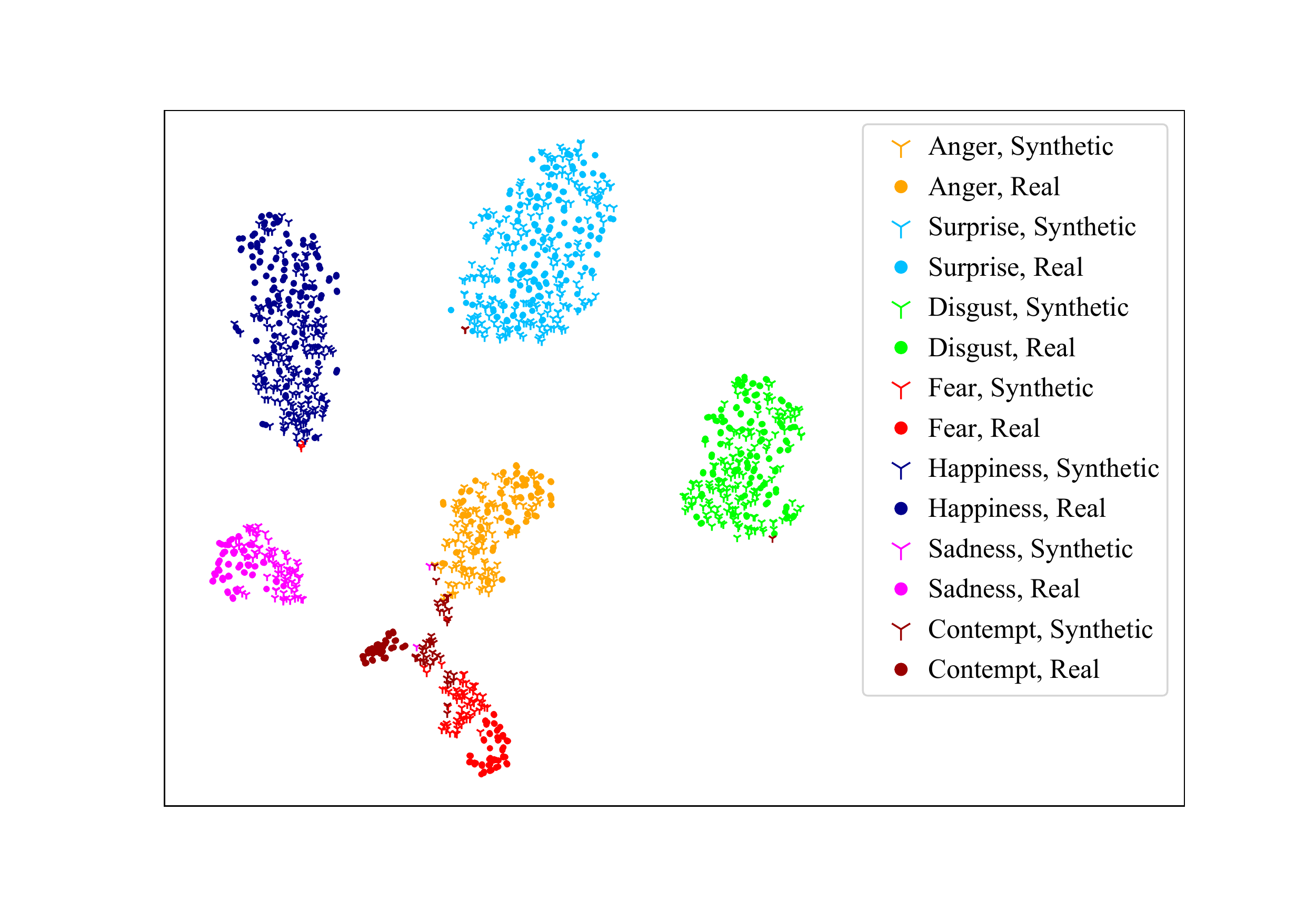}%
\label{fig:with_IL_allexp}}
\hfil
\subfloat[]{\includegraphics[width=1.7in]{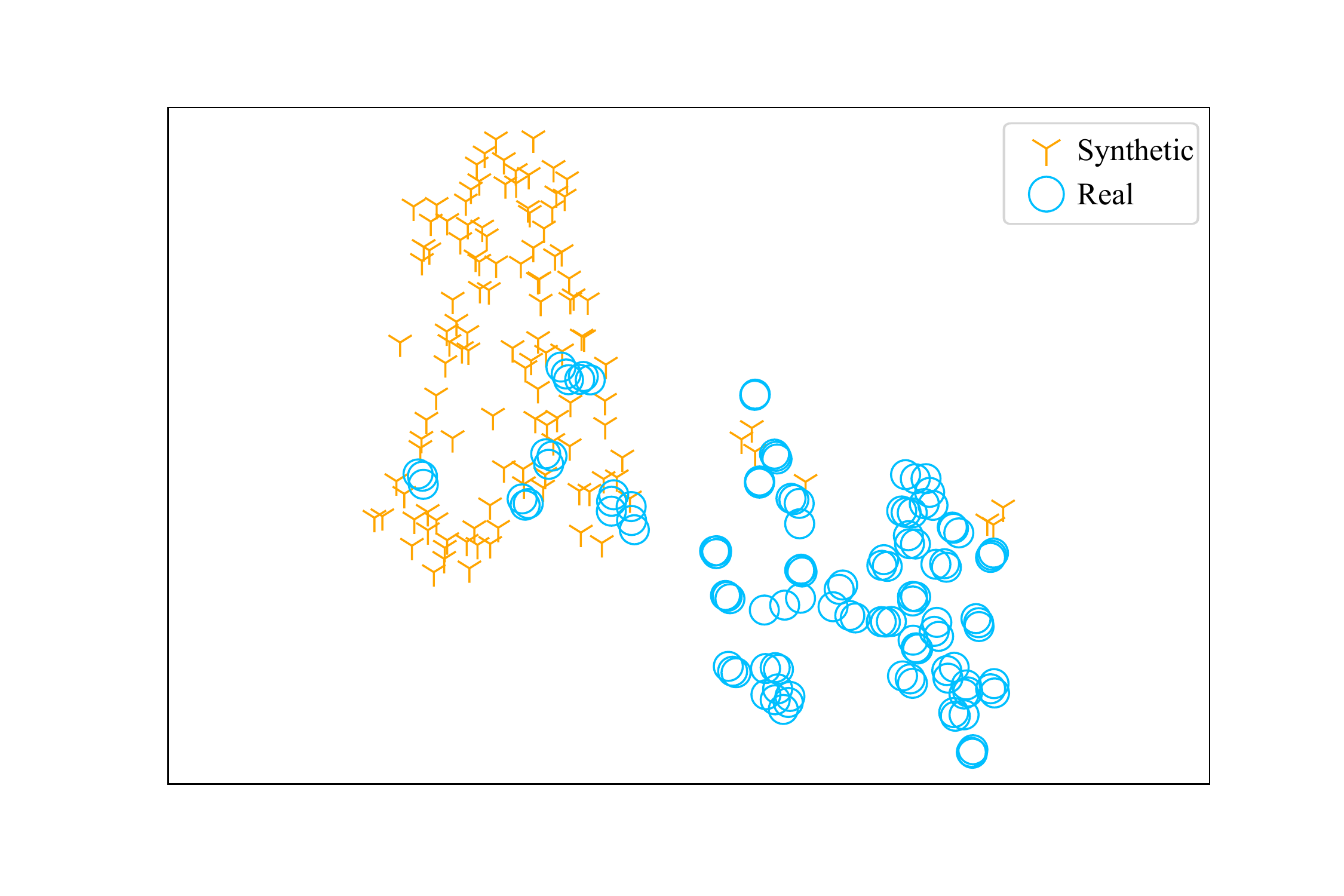}%
\label{fig:without_IL_anger}}
\subfloat[]{\includegraphics[width=1.7in]{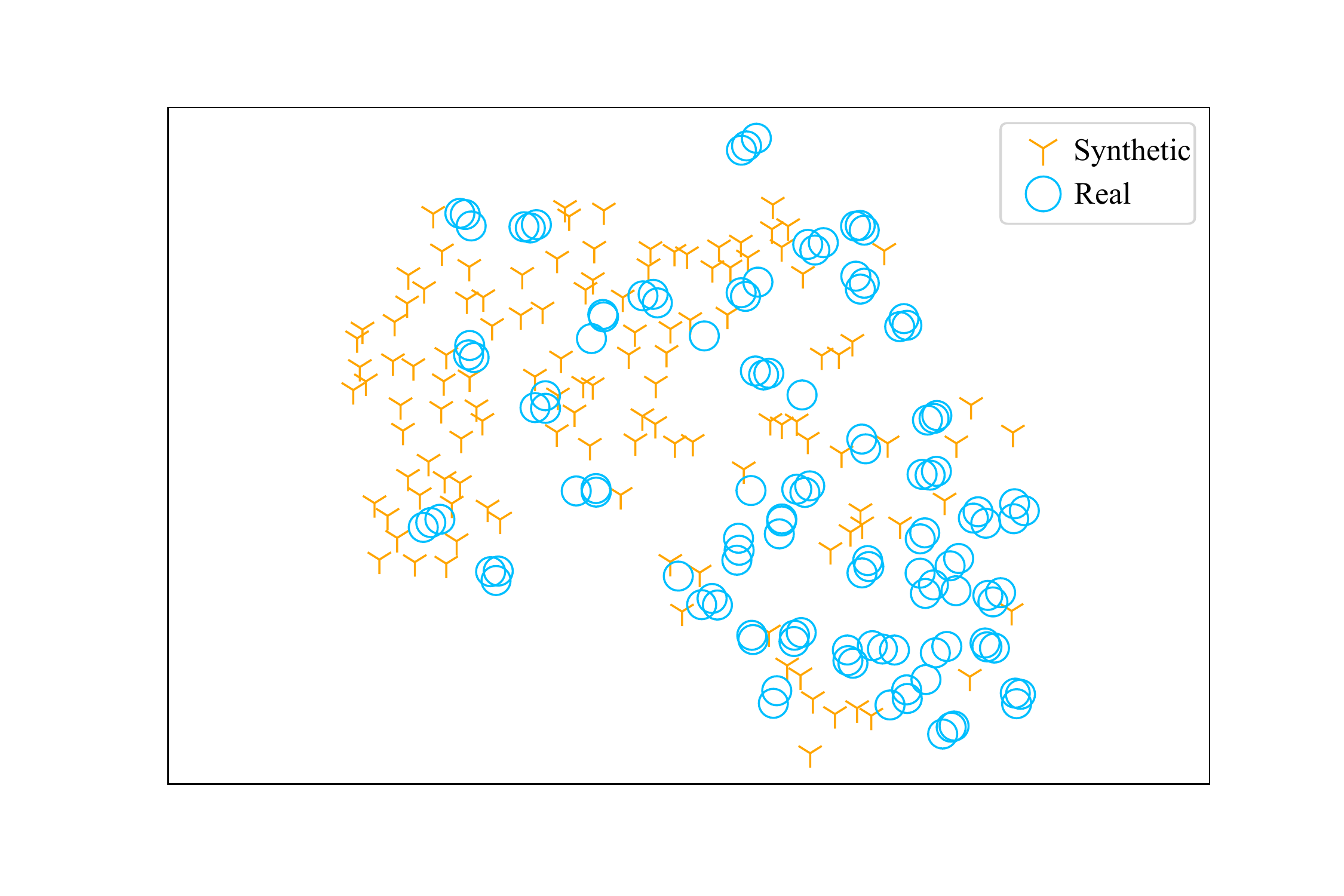}%
\label{fig:with_IL_anger}}
\caption{\small Visualization of the features extracted from different models, which are trained on the CK+ database. (a) Feature visualization of all expressions on the model trained by FESR\_JL-IL. (b) Feature visualization of all expressions on the model trained by FESR\_JL. (c) Feature visualization of the anger expression on the model trained by FESR\_JL-IL. (d) Feature visualization of the anger expression on the model trained by FESR\_JL.}
\label{fig:vis}
\end{figure}

To demonstrate the effectiveness of the intra-class loss on the final features for FER, we visualize the features by t-SNE \cite{maaten2008visualizing}. Fig. \ref{fig:vis} shows the feature visualization of the real images and the synthetic images obtained by FESR\_JL-IL and FESR\_JL, respectively. In Fig.~9(a) and Fig.~9(c), the feature distributions of the real images and the synthetic images are separated. However, in Fig.~9(b) and Fig.~9(d), by taking advantaging of the intra-class loss, these two distributions become close to each other. This indicates that minimizing the intra-class loss can significantly mitigate the problem of data bias between the real images and the synthetic images.

\subsubsection{The effectiveness of RDBP}
\label{sec:effect_RDBP}

The performance obtained by FESR\_JL-RDBP (using the standard SGD to minimize the intra-class loss between the real images and the synthetic images) is worse than that obtained by FESR\_JL (using the real data-guided back-propagation, i.e., RDBP, to minimize the intra-class loss). This is because that the traditional SGD method equally treats the features of real images and synthetic images. Thus, it encourages the features of images to get close to each other when minimizing the intra-class loss. As a result, the interference of the synthetic images can decrease the discriminability of the features of real images. On the contrary, RDBP regards the features of real images as the ground-truth to supervise the features of synthetic images (by ignoring the gradients computed from the real images). In other words, the feature distribution of synthetic images is enforced to approach that of real images while avoiding the inverse (i.e.,
the feature distribution of real images is unexpectedly
enforced to get close to that of synthetic images, thus decreasing the discriminative ability of the features). Therefore, by using RDBP, the parameters of the recognition network $R$ can be optimized for better performance.

\subsubsection{The effectiveness of the synthetic images with new identities}

Compared with FESR\_JL-IL, FESR\_Real obtains the similar accuracy rate on the CK+ database. However, on the more challenging Oulu-CASIA and MMI databases, the accuracy rates obtained by FESR\_Real have obvious decrease (1.75\% and 1.3\%, respectively), which demonstrates the effectiveness of the synthetic images with new identities. This is mainly due to the fact that the synthetic images generated by FESR\_Real have the same identities as the input images. In contrast, FESR\_JL-IL can generate synthetic images with new identities (not existing in the training set). Therefore, the auxiliary training set used by FESR\_JL-IL has more diversity than that used by FESR\_Real, which is helpful for improving the final performance of FER.


\subsubsection{The effectiveness of two-stage learning procedure}
Compared with  FESR\_OneSt (one-stage learning), FESR\_JL (two-stage learning)  obtains much higher recognition rates on all the databases. Therefore, when FESGAN and the recognition network $R$ are jointly trained
from the beginning (i.e., one-stage learning), the performance of our method significantly drops.
This is mainly because that the image quality
of the generated images is poor and the recognition network $R$
may be seriously affected due to the interference of these
generated images. Moreover, the classification loss from $R$ can dominate the capacity of facial expression modelling of
the generator in FESGAN. As a result, the generator will cater
to the recognition network $R$ and thus the diversity and recognition difficulty of generated facial images may be reduced. In contrast,
the two-stage learning procedure alleviates the above problem by firstly pre-training  FESGAN to generate facial images with different
facial expressions and then performing  joint learning of FESGAN and $R$. In this way,
$R$ will be fed with
photo-realistic generated images in the second stage, and thus the generator of FESGAN will not be greatly influenced by
$R$ (it needs to balance the loss from the generator, two discriminators and the recognition network). As a result, the generator is able to generate images with high-diversity and these images are useful for improving the performance of $R$.
%
%

\subsection{Comparisons with the State-of-the-art Methods}
\label{sec:cmp_state}

In this subsection, we compare the proposed method with several state-of-the-art methods, including both traditional methods (LBP-TOP \cite{zhao2007dynamic} 
and STM-ExpLet \cite{liu2014learning}
) and the recently-proposed deep learning-based methods, including 3D-CNN-DAP \cite{liu2014deeply}, AUDN \cite{liu2015inspired}, IACNN \cite{meng2017identity}, DTAGN \cite{jung2015joint}, PHRNN-MSCNN \cite{zhang2017facial}, ExprGAN+R \cite{ding2018exprgan} and DeRL \cite{yang2018facial}. Here, ExprGAN-R denotes the method which directly leverages the publicly released model of ExprGAN  to generate images with all expressions from the input image, and then uses these generated images as the auxiliary training data to train the recognition network $R$.
The CK+, Oulu-CASIA, MMI databases are employed.

Table \ref{tab:per_comparison} shows the recognition performance obtained by all the competing methods on the three databases. The proposed method achieves the top performance among all the methods. This is because the joint learning strategy and the intra-class loss used in the proposed method. We can see that all the methods achieve good recognition rates on the CK+ database, since all the face images in CK+ are captured in the frontal view and have obvious facial expressions variations. However, the recognition rates on the MMI database are much worse than those in the other two databases, which demonstrates the significant challenge of MMI. Both FESR\_SL and ExprGAN+R perform facial expression synthesis and facial expression recognition independently. However, FESR\_SL achieves better performance than ExprGAN+R, which demonstrates that the proposed FESGAN can synthesize facial images beneficial for facial expression recognition.

We can observe that the top 3 methods (i.e., the proposed FESR\_JL, PHRNN-MSCNN, and DeRL) on the CK+ and Oulu-CASIA databases and the top 2 methods (i.e., the proposed FESR\_JL and PHRNN-MSCNN) on the MMI database all belong to the deep learning-based methods, which further demonstrates the powerful ability of deep learning.
Specifically, DeRL uses cGAN to generate neutral face images having the same identity as the input. The intermediate layers of the generative model encode the expressive information, which can be used to perform person-independent FER. PHRNN-MSCNN utilizes a CNN and a recurrent neural network (RNN) to respectively extract spatial features (based on still frames of the facial image sequences) and temporal features (based on facial landmarks sequences). Furthermore, an expression verification signal is employed to reduce the within-expression variations to boost the performance for FER.
\begin{table}[!t]
\renewcommand{\arraystretch}{1.1}
\caption{\small Recognition rates (\%) obtained by the different methods on the CK+, Oulu-CASIA and MMI databases. ``-'' indicates that the corresponding results are not provided by the method.  The best results are boldfaced.}
\label{tab:per_comparison}
\centering
\scalebox{1.1}{
\begin{tabular}{|c|c|c|c|} 
\hline
\multirow{2}{*}{Methods} & \multicolumn{3}{c|}{Accuracy}\\
\cline{2-4} & CK+ & Oulu-CASIA & MMI \\
\hline
LBP-TOP \cite{zhao2007dynamic} & 88.99 & 68.13 & 59.51\\
3D-CNN-DAP \cite{liu2014deeply} & 92.40 & - & 63.40\\
AUDN \cite{liu2015inspired} & 93.70 & - & 75.85 \\
STM-ExpLet \cite{liu2014learning} & 94.19 & 74.59 & 75.12\\
IACNN \cite{meng2017identity} & 95.37 & - & 71.55\\
DTAGN \cite{jung2015joint} & 97.25 & 81.46 & 70.24\\
ExprGAN+R & - & 83.75 & -\\
DeRL \cite{yang2018facial} & 97.30 & 88.00 & 73.23\\
PHRNN-MSCNN \cite{zhang2017facial} & 98.50 & 86.25 & 81.18 \\
\hline
BASELINE & 95.63 & 82.92 & 79.88 \\
FESR\_SL & 96.82 & 84.17 & 81.66 \\
FESR\_JL-IL & 97.18 & 85.21 & 82.80 \\
FESR\_JL & \textbf{99.34} & \textbf{88.13} & \textbf{84.81} \\
\hline
\end{tabular}}
\end{table}

Both DeRL and PHRNN-MSCNN achieve the outstanding performance by elaborately designing the network architecture or the objective functions for FER. However, in this paper, we train a recognition network with a simple architecture for FER, where a pre-trained CNN (i.e., a part of the Light CNN-29 trained on the large database for face recognition \cite{wu2018light}) is used to extract the low-level features of the images. In this manner, the risk of overfitting can be significantly reduced during the training of the recognition network. In summary, the experimental results show the superiority of the proposed method for FER in comparison with other methods.

Note that the network architecture of FESGAN is relatively simple (see Table I for more detail). Both the generator and discriminators only consist of very few convolutional layers and FC layers.
Therefore, the training of these networks is sufficient by using a small amount of training samples. In fact,  recent state-of-the-art methods, such as ExprGAN \cite{ding2018exprgan}, G2-GAN \cite{qiao2018geometry}, have also shown the impressive synthesis performance on small databases (such as CK+, Oulu-CASIA and MMI). For the recognition network $R$, we adopt the strategy of pre-training and fine-tuning. Specifically, the recognition network is composed of a part
of the Light CNN-29 (trained from the large-scale CASIA-WebFace dataset \cite{Yi2014} and MS-Celeb-1M dataset \cite{Guo2016}), two convolutional layers, two
fully-connected layers and a fully-connected layer. The weights of the layers from the Light CNN-29 are frozen all the time, while the following layers are fined-tuned during
the joint training process.
Based on the elaborately designed network architecture, we perform joint learning of two related tasks (i.e., facial expression synthesis and recognition), where the training data can be effectively augmented to improve the performance of expression recognition. Moreover, we propose an intra-class loss to implicitly reduce the intra-class variations between the real images and the synthetic ones.
Therefore, our proposed method can achieve superior performance by only using a small amount of training samples.

\subsection{Comparisons on Other Databases}
\label{otherdatasets}
We also evaluate the proposed method on the large-scale databases, including Multi-PIE and TFD databases. We compare the proposed method with several state-of-the-art methods, including kNN \cite{Bishop2006}, LDA \cite{Bishop2006}, LPP \cite{He2004}, D-GPLVM \cite{Urtasun2007}, GPLRF \cite{Zhong2010}, GMLDA \cite{Sharma2012}, GMLPP \cite{Sharma2012}, MvDA \cite{Kan2012},  DS-GPLVM \cite{Eleftheriadis2015}, Gabor+PCA \cite{Dailey2002}, Deep mPoT \cite{Ranzato2011}, CDA+CCA  \cite{Rifai2012}, disRBM \cite{Reed2014}, bootstrap-recon \cite{Reed2014b}, VGG \cite{Simonyan2015} and FN2EN \cite{Ding2017}.
Tables \ref{tab:PIE} and \ref{tab:TFD} show the recognition accuracy obtained by all the methods on the Multi-PIE and TFD databases, respectively. Note that the recognition rates in Tables \ref{tab:PIE} and \ref{tab:TFD} obtained by the competing methods are taken from \cite{Eleftheriadis2015} and \cite{Ding2017}, respectively.

From Tables \ref{tab:PIE} and \ref{tab:TFD}, we can see that our proposed method outperforms all the other competing methods by about 2\% to 15\% improvement in terms of recognition rate, which further validates the effectiveness of the proposed method on large-scale databases.
\begin{table}[!t]
\renewcommand{\arraystretch}{1.1}
\caption{\small Recognition rates (\%) obtained by the different methods on the Multi-PIE database. The best results are boldfaced.}
\label{tab:PIE}
\centering
\scalebox{0.98}{
\begin{tabular}{|c|c|c|c|c|c|c|} 
\hline
\multirow{2}{*}{Methods} & \multicolumn{5}{c|}{Pose} &\multirow{2}{*}{Average}\\
\cline{2-6}          & \multicolumn{1}{c|}{$-30^{\circ}$} & \multicolumn{1}{c|}{$-15^{\circ}$} & \multicolumn{1}{c|}{$0^{\circ}$} & \multicolumn{1}{c|}{$15^{\circ}$} & \multicolumn{1}{c|}{$30^{\circ}$} &  \\
\hline
kNN  &80.88   & 81.74   & 68.36  &  75.03  & 74.78 & 76.15   \\
LDA \cite{Bishop2006} &92.52    & 94.37  & 77.21  & 87.07  & 87.47  & 87.72  \\
LPP \cite{He2004} & 92.42    & 94.56   & 77.33  & 87.06   & 87.68 & 87.81  \\
D-GPLVM \cite{Urtasun2007} &91.65    & 93.51  &78.70   &85.96   &86.04  &87.17  \\
GPLRF \cite{Zhong2010} & 91.65    &93.77   &77.59   &85.66   &86.01  &86.93  \\
GMLDA \cite{Sharma2012} & 90.47   & 94.18 & 76.60  & 86.64  & 85.72 & 86.72 \\
GMLPP \cite{Sharma2012}  & 91.86    &94.13   & 78.16  & 87.22  & 87.36 &87.74\\
MvDA \cite{Kan2012}   & 92.49    & 94.22   & 77.51  & 87.10  &87.89  &87.84\\
DS-GPLVM \cite{Eleftheriadis2015} & \textbf{93.55}  &\textbf{96.96}   & 82.42   &89.97   &90.11  &90.60 \\
\hline
BASELINE & 89.75   &  90.12 &  83.01  &88.97   & 88.10 &  87.99 \\
FESR\_JL & 92.28   &  94.23 &  \textbf{86.94}  &\textbf{92.29}   & \textbf{94.71}  &\textbf{92.09} \\
\hline
\end{tabular}}
\end{table}

\begin{table}[!t]
\renewcommand{\arraystretch}{1.1}
\caption{\small Recognition rates (\%) obtained by the different methods on the TFD database. The best results are boldfaced.}
\label{tab:TFD}
\centering
\scalebox{1.1}{
\begin{tabular}{|c|c|} 
\hline
{Methods} & {Accuracy}\\
\hline
Gabor+PCA \cite{Dailey2002} & 80.20\\
Deep mPoT \cite{Ranzato2011} & 82.40\\
CDA+CCA \cite{Rifai2012}  &  85.50\\
disRBM \cite{Reed2014} & 85.40\\
bootstrap-recon \cite{Reed2014b}  & 86.80\\
VGG  \cite{Simonyan2015} & 86.70\\
FN2EN \cite{Ding2017} & 88.90\\
\hline
BASELINE &85.45\\
FESR\_JL & \textbf{90.12}\\
\hline
\end{tabular}}
\end{table}
\section{Conclusion}

In this paper, we propose a novel method that integrates facial expression synthesis and recognition in a unified framework for effective FER. In particular, the classification loss computed from the recognition network for the synthetic images is used to supervise the learning of both FESGAN and the recognition network. In this way, the generator of FESGAN is encouraged to generate synthetic images beneficial to the recognition network. We elaborately design FESGAN to generate images with new identities (by replacing the latent face representation with random vectors sampled from a prior distribution) to increase the diversity of the training data. An intra-loss with a novel real data-guided back-propagation (RDBP) algorithm is introduced to mitigate the problem of data bias between the real images and the synthetic images. Such a strategy not only significantly reduces the intra-class variations of the features, but also effectively encourages the recognition network to make full use of the information in both real images and synthetic images for FER. Extensive experiments have been conducted on public facial expression databases and the experimental results have demonstrated the superior performance of the proposed method over several state-of-the-art methods.

\section*{Acknowledgements}
\small This work was supported by the National Key R\&D Program of China under Grant 2017YFB1302400, by the National Natural Science Foundation of China under Grants 61571379, U1605252, 61872307, by the Natural Science Foundation of Fujian Province of China under Grants 2017J01127 and 2018J01576.


%





\ifCLASSOPTIONcaptionsoff
  \newpage
\fi


\normalem
\bibliography{IEEEabrv,reference}

\end{document}